\documentclass[10pt,twocolumn,letterpaper]{article}

\usepackage{cvpr}              

\usepackage[dvipsnames]{xcolor}

\definecolor{cvprblue}{rgb}{0.21,0.49,0.74}
\usepackage[pagebackref,breaklinks,colorlinks,linkcolor=cvprblue,urlcolor=cvprblue,citecolor=cvprblue]{hyperref}

\usepackage{amsmath}
\usepackage{amssymb}
\usepackage{mathtools}
\usepackage{graphicx}
\usepackage{caption}
\usepackage{booktabs} 
\usepackage{multirow} 
\usepackage{adjustbox}
\usepackage{tabularx}
\usepackage{makecell}
\newcolumntype{C}[1]{>{\centering\arraybackslash}m{#1}}

\def\paperID{6401} 
\def\confName{CVPR}
\def\confYear{2024}

\newcommand{\modelname}{DN2N}

\title{Editing 3D Scenes via Text Prompts without Retraining}

\author{
Shuangkang Fang$^{1}$
\qquad
Yufeng Wang$^1$
\qquad
Yi Yang$^2$\\[2pt]
\qquad
Yi-Hsuan Tsai$^3$
\qquad
Wenrui Ding$^1$
\qquad
Shuchang Zhou$^2$
\qquad
Ming-Hsuan Yang$^{345}$
\\[9pt]
$^1$\tt Beihang University \qquad $^2$Megvii  \qquad $^3$Google  \\[2pt]
\tt \qquad $^4$University of California, Merced \qquad $^5$Yonsei University
}

\begin{document}

\twocolumn[{%
\renewcommand\twocolumn[1][]{#1}%
\maketitle
\begin{center}
    \centering
    \captionsetup{type=figure}
    \includegraphics[width=1\textwidth]{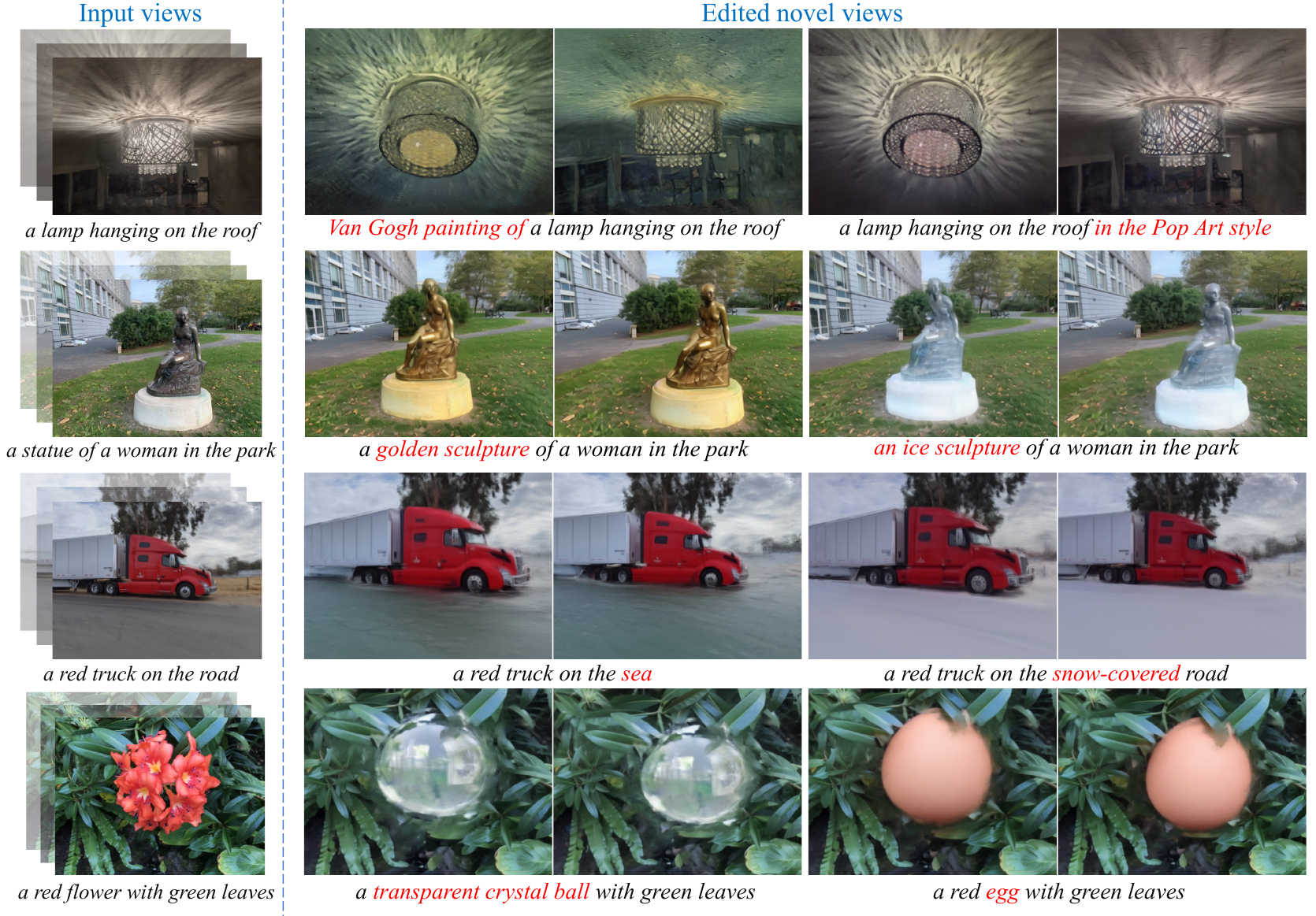}
\caption{\textbf{Visualization results of our method in 3D scene editing}. The proposed \modelname{} method enables users to obtain realistic and 3D consistent novel views that correspond to the target caption without retraining a new model. 
}
        \label{fig:shocking}
\end{center}%
}]

\begin{abstract}
Numerous diffusion models have recently been applied to image synthesis and editing. 
However, editing 3D scenes is still in its early stages. It poses various challenges, such as the requirement to design specific methods for different editing types, retraining new models for various 3D scenes, and the absence of convenient human interaction during editing.
To tackle these issues, we propose a novel text-driven 3D editing method with generalization capability,
termed \modelname{}, which allows for the \textit{direct} acquisition of the editing results without the requirement for retraining. 
Our method employs off-the-shelf text-based editing models of 2D images to modify the 3D scene images, followed by a filtering process to discard poorly edited images that disrupt 3D consistency. 
We then consider the remaining inconsistency as a problem of removing noise perturbations, which can be solved by generating data with similar perturbation characteristics for training.
We then propose cross-view regularization terms to help the DN2N model mitigate these perturbations.
Empirical results show that our method achieves multiple editing types, including but not limited to appearance editing, weather transition, object changing, and style transfer. 
Most significantly, our method exhibits strong generalization of editing capabilities, eliminating the need to customize or retrain editing models for specific scenes or editing types. Therefore, it significantly reduces editing time and storage consumption. The project website is available at \emph{\url{https://sk-fun.fun/DN2N}}.
\end{abstract}
    
\section{Introduction}
Significant advances in neural radiance fields (NeRF) techniques~\cite{mildenhall2020nerf, yu2021plenoctrees,muller2022instant,barron2021mip,zhang2020nerf++,wang2021ibrnet,lin2021barf,chen2022tensorf} have been modeled for
specific types of 3D manipulation, such as appearance editing~\cite{martin2021nerfinthewild, kobayashi2022decomposing}, scene composition~\cite{tang2022CCNeRF, tancik2022block}, weather transformation~\cite{li2022climatenerf}, multiple editing~\cite{fang2022pvd,fang2023pvdal}, and style transfer~\cite{huang2022stylizednerf, gu2021stylenerf,nguyen2022snerf-miss1,fan2022unified-miss2,zhang2022arf}. 
Recently, a few attempts have been made to leverage multimodal techniques to design text-guided 3D editing methods~\cite{wang2022clipnerf,wang2022nerfart, haque2023instruct}. 
Despite the demonstrated success, several challenges remain: (1) many of these techniques~\cite{martin2021nerfinthewild, kobayashi2022decomposing,tang2022CCNeRF, tancik2022block,li2022climatenerf,fang2022pvd,fang2023pvdal,huang2022stylizednerf, gu2021stylenerf,nguyen2022snerf-miss1,fan2022unified-miss2,zhang2022arf} are less user-friendly; (2) existing methods~\cite{martin2021nerfinthewild, kobayashi2022decomposing,tang2022CCNeRF, tancik2022block,li2022climatenerf,fang2022pvd,fang2023pvdal,huang2022stylizednerf, gu2021stylenerf,nguyen2022snerf-miss1,fan2022unified-miss2,zhang2022arf} typically rely on known editing types in advance, resulting in limited modification capabilities; 
(3) retraining an editing model is required~\cite{martin2021nerfinthewild, kobayashi2022decomposing,tang2022CCNeRF, tancik2022block,li2022climatenerf,fang2022pvd,fang2023pvdal,huang2022stylizednerf, gu2021stylenerf,nguyen2022snerf-miss1,fan2022unified-miss2,zhang2022arf,wang2022clipnerf,wang2022nerfart, haque2023instruct} for each particular 3D scene, leading to computational and memory overhead. 
To tackle these challenges, in this article, we aim to devise a novel approach that possesses diverse and user-friendly editing capabilities while also having the ability to directly edit new scenes without retraining.

Currently, leveraging existing multimodal models on 2D images~\cite{radford2021clip,hertz2022prompt,brooks2022instructpix2pix,mokady2022null} and NeRF-based framework allows for text-driven 3D scene editing~\cite{wang2022clipnerf,wang2022nerfart, haque2023instruct}.
However, these methods are non-generalizable since if we wanted to perform 100 types of editing on 100 scenes, they would require training 10,000 models. In contrast, our research aims to accomplish these edits with a single model.
An intuitive solution is to replace the NeRF model in these methods with a generalizable one~\cite{wang2021ibrnet,chen2021mvsnerf,liu2022neuray,johari2022geonerf}. 
Nevertheless, directly designing it in this way presents two issues: (1) insufficient task-specific training data to ensure the model's robust generalization abilities; (2) the inherent 3D inconsistency arising from directly using multimodal 2D image editing models to edit the 3D scenes.
To address these issues, we propose directly modeling the inherent 3D inconsistency, and leveraging existing tools to generate sufficient training data with similar characteristics, enabling the model to acquire robust generalization abilities to mitigate this inconsistency.

Specifically, we initially utilize a 2D editing model~\cite{mokady2022null} to perform the preliminary editing on the images of a 3D scene. We subsequently apply a designed content filter to remove several images with poor editing results.
The remaining images may still contain inconsistent 3D results, which we consider as minor perturbations to the ideally edited images due to the inherent stochastic and diverse nature of the 2D editing model.
To remove such minor perturbations, we generate ample amounts of training data pairs with similar characteristics to train our model: firstly, we utilize the BLIP model~\cite{li2023blip2} and GPT~\cite{brown2020gpt3} to obtain the input and target captions required for editing, then apply random slight edits to the 3D scene images to simulate the minor perturbations.
In this manner, images with added random perturbations are utilized as inputs, while clean images serve as ground truth for training the generalizable model.

We further introduce two cross-view regularization terms during training to facilitate its 3D consistency, including the self and neighboring views. 
The former requires the model to generate consistent results for the same target view that is derived from two different source views, while the latter enforces the overlapping pixel values between the target and adjacent views to be approximately close.

The main contributions of this work are:
\begin{itemize}[nosep,leftmargin=*] 
\item We develop a 3D scene editing framework with generalization capabilities, named \modelname{}. It employs off-the-shelf 2D editing models for 3D scene manipulation, where the induced 3D inconsistency is modeled as perturbations and addressed by generating training data pairs with similar perturbation characteristics for optimization.
\item We devise a generalizable NeRF model structure. In addition to crafting sufficient training data pairs, we integrate cross-view regularization terms into the model's training process, enabling it to acquire the capability to eliminate perturbations.
\item We conduct extensive experiments of different editing types on multiple datasets. 
Compared with other approaches, DN2N offers diverse editing capabilities within a shorter time and lower memory overhead, as well as eliminating the necessity for customizing or retraining a model for different scenes or editing types.
\end{itemize}
\section{Related work}

\begin{figure*}[htbp]
    \centering
    \begin{subfigure}{1\textwidth}
        \centering
        \includegraphics[width=\textwidth]{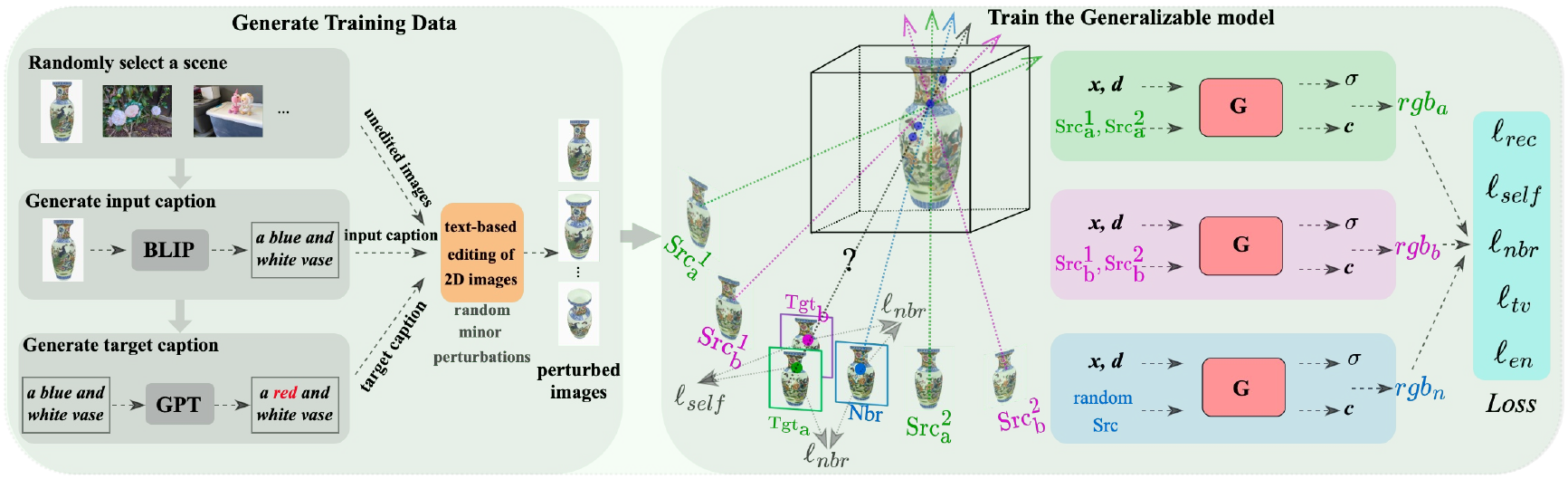}
        \caption{\textbf{Training stage}. \textit{Left}: we generate abundant training data pairs by applying subtle perturbations to all training scenes using the BLIP, GPT, and a 2D editing model. 
        \textit{Right}: we train the generalized NeRF model $G$ by incorporating cross-view regularization terms $\ell_{self}$ and $\ell_{nbr}$. Upon completion of the training process, there is no need to retrain the model for new scenes. Instead, the editing results can be directly obtained through the inference stage.}
        \label{fig:subfigure1}
        \vspace{0.3cm}
    \end{subfigure}
    \vspace{0.2cm} %
    \begin{subfigure}{1\textwidth}
        \centering
        \includegraphics[width=\textwidth]{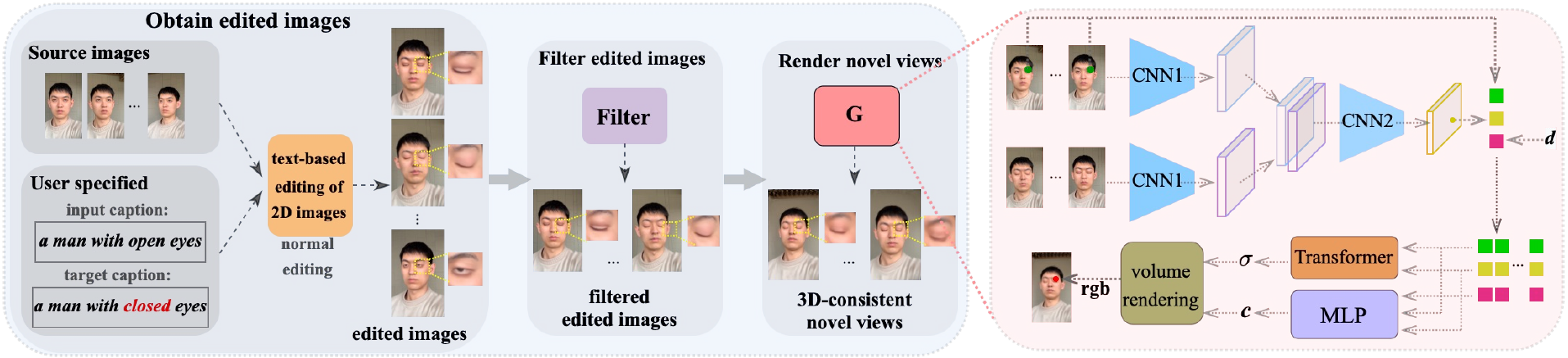}
        \caption{\textbf{Inference stage}. \textit{Left}: we begin by applying standard magnitude editing to the 3D scene images. \textit{Middle}: we devise a content filter to eliminate the images with subpar editing results and compromised consistency. \textit{Right}: then, we utilize the well-trained model $G$ to obtain the edited novel views.}
        \label{fig:subfigure2}
    \end{subfigure}
    \caption{\textbf{Illustration of the \modelname{} framework}. See Sec~\ref{sec:method} for the detailed description.}
    \label{fig:method_pipeline}
\end{figure*}

\textbf{NeRF-based novel view synthesis}. 
Novel view synthesis based on NeRF has recently gained significant attention in the vision and graphics communities~\citep{mildenhall2020nerf, yu2021plenoctrees,muller2022instant,barron2021mip,zhang2020nerf++,wang2021ibrnet,lin2021barf,chen2022tensorf}. 
NeRF represents the structure and appearance of a 3D scene by using a neural network that takes the spatial location and view direction as input and outputs the corresponding color and opacity at each pixel.
Subsequent works have improved NeRF, such as speeding up the training process~\citep{muller2022instant, chen2022tensorf}, designing better sampling strategies~\citep{barron2021mip}, and enhancing generalization ability~\citep{wang2021ibrnet, johari2022geonerf,liu2022neuray,chen2021mvsnerf}.

\noindent \textbf{Diffusion-based image editing}.
Diffusion-based models have been widely used in image generation~\citep{sohl2015deep,dhariwal2021diffusion,ho2020denoising,saharia2022photorealistic,ramesh2022hierarchical,rombach2022high,song2020denoising, song2020score}. 
Numerous text-based image editing methods have recently been developed, such as GLIDE~\citep{nichol2021glide} and Stable Diffusion~\citep{rombach2022high}. 
Imagic~\citep{kawar2022imagic} finetunes images according to text, and 
Prompt-to-prompt~\citep{hertz2022prompt} preserves unedited regions by utilizing cross-attention information. 
Pix2pix-zero~\citep{parmar2023zero} employs embedding vector mechanisms to establish controllable editing directions for images. 
InstructPix2Pix~\citep{brooks2022instructpix2pix} trains a model by generating a large number of text-editing image pairs using GPT~\citep{brown2020languageGPT} and Stable Diffusion. 
Null-text inversion~\citep{mokady2022null} proposes an accurate inversion process to enhance image-controlled editing capabilities.

\noindent \textbf{3D scene editing}.
Numerous 3D scene editing approaches have been developed based on point cloud~\citep{huang2021learning,mu20223d} and triangle meshes representations~\citep{zhou2014color,hollein2022stylemesh,han2021exemplar}. 
These methods are limited by their inherent representations, which restrict scalability and editing capability. 
Recently, NeRF-based methods have been used for 3D scene editing~\cite{martin2021nerfinthewild, kobayashi2022decomposing,tang2022CCNeRF, tancik2022block,li2022climatenerf,fang2022pvd,fang2023pvdal,huang2022stylizednerf, gu2021stylenerf,nguyen2022snerf-miss1,fan2022unified-miss2,zhang2022arf,boss2021nerd,boss2021neural,mu20223dphotostylization}.
However, several aspects of these methods are limited. 
For instance, they are usually restricted to performing a single editing type.
Several methods have overcome this limitation~\cite{wang2022clipnerf,wang2022nerfart, haque2023instruct}. For example,
%
Instruct-N2N~\citep{haque2023instruct} enables diverse controllable editing of 3D scenes by pre-editing 2D images using InstructPix2Pix~\citep{brooks2022instructpix2pix}.
However, Instruct-N2N also has two notable drawbacks. 
First, retraining is necessary for each new editing direction, resulting in significant computation and memory overhead.
Second, this method requires continuously generating new training data and updating model parameters during training, which is time-consuming and has high storage consumption. 
In contrast, our method addresses these two challenges by only using a single editing model for all 3D scenes and editing types without the need for retraining, resulting in decreased model storage consumption and training time.

\section{Method} \label{sec:method}

\textbf{Overall framework}. The DN2N framework is illustrated in Figure~\ref{fig:method_pipeline}.
The training stage involves training the model across multiple scenes. For each scene,
we first utilize BLIP~\cite{li2023blip2} to obtain its description as the input caption and then employ GPT~\cite{brown2020gpt3} to randomly modify this caption to create the target caption.
We then use a 2D image editing model~\cite{mokady2022null} to apply minor perturbations to the scene based on these captions to obtain training data pairs. The training objective of our model is to remove these perturbations and reconstruct the 3D scene (see Eq.~\ref{eq:argmin_real}). 
Besides, we introduce cross-view regularization terms to assist in model training. Specifically, we use two sets of independent source views to render the same target view, resulting in $Tgt_a$ and $Tgt_b$, respectively, and render a neighboring view around the target view to obtain $Nbr$. Then we impose a consistency loss (see Eq.~\ref{eq:self_view} and Eq.~\ref{eq:neigh}) on the three rendered results. 
During the inference stage, given a new scene, we begin by applying standard magnitude editing to the scene and filter out images with poor editing effects (see Eq.~\ref{eq:filter}).
Finally, we feed the filtered images into the well-trained generalizable model to obtain edited novel views directly.

\noindent \textbf{Optimization objectives}.
A 3D scene training data consists of $N$ images and their corresponding camera parameters, denoted as $\{I_{i}, P_i\}_{i=1}^{N}$. 
We employ the 2D image editing model $\mathcal{F}$ to obtain the pre-edited image $\tilde{I}_{i}$ for each $I_i$, which is defined as:
\\[-2pt]
\begin{equation}
\tilde{I}_{i} = \mathcal{F}_{}(I_i, C_{in}^i, C_{tgt}^i, \theta), 
\label{eq:gen_edit}
\end{equation}
\\[-3pt]
where $C_{in}^i$ is input caption for the unedited image $I_i$, $C_{tgt}^i$ is target caption for the edited image $\tilde{I}_{i}$, and $\theta$ is the hyper-parameters of model $\mathcal{F}$.
After filtering out the poorly edited images in $\{\tilde{I}_{i}, P_i\}_{i=1}^{N}$, the resulting data is denoted as $\{\tilde I_m, P_m\}_{m=1}^M$, where $M \le N$. Then the generalizable NeRF model $G$ with parameter $\Theta$ predicts the target view $\hat{I}_{m}^{}$ using $K$ source views $\{\tilde I_k, P_k\}_{k=1}^K$:
\\[-2pt]
\begin{equation}
\hat{I}_{m}^{} = G(\tilde I_{k}, P_k, \Theta \mid k=1,\dots,K, P_k \ne P_m).
\end{equation}
\\[-3pt]
Assuming that the ideal ground truth (GT) of edited images with 3D consistency is denoted as ${I}^{gt}_{m}$, our optimization objective can be expressed as:
\\[-2pt]
\begin{equation}
\arg\min_{\Theta}\sum_{m=1}^{M}\left \| {I}^{gt}_{m} - \hat{I}_{m}   \right \|^2.
\label{eq:argmin_ori}
\end{equation}
\\[-2pt]
However, the GT is not at our disposal.  
Thus, we express the pre-edited image $\tilde{I}_{m}$ as $\tilde{I}_{m} = {I}^{gt}_{m} + \bigtriangleup I_m$.
This implies that the inconsistent image $\tilde{I}_{m}$ can be perceived as consistent GT image ${I}^{gt}_{m}$ with minor perturbations $\bigtriangleup I_m$. 
If our model $G$ can learn the ability to remove $\bigtriangleup I_m$ from the $\tilde{I}_{m}$, then we can obtain the consistent editing results since ${I}^{gt}_{m}$=$\tilde{I}_{m}-\bigtriangleup I_m$.
In this work, the perturbations $\bigtriangleup I_m$ mainly stem from the inherent editing characteristics of model $\mathcal{F}$.
Hence, we apply similar random minor perturbations ${\bigtriangleup I_{i}}$ to clean 3D scene images 
$\{I_{i}\}_{i=1}^{N}$  by controlling the parameters $\theta$ in $\mathcal{F}$ to get the training data $\{\tilde I_{i}, P_i\}_{i=1}^{N}$.
Then, the unedited images $\{I_{i}\}_{i=1}^{N}$ can be used as pseudo ground truth to train the model $G$ as follows:
\begin{equation}
\resizebox{.9\hsize}{!}{$\arg\min_{\Theta}\sum_{i=1}^{N}\left \| I_i \!-\! G(\tilde I_k, P_k, \Theta \mid k\!=\!1,\dots,K, P_k \!\ne\!P_i)   \right \|^2.$}
\label{eq:argmin_real}
\end{equation}

\noindent  \textbf{Generate training data}.
To generate training data pairs, we utilize the off-the-shelf tools BLIP~\cite{li2023blip2} and GPT~\cite{brown2020gpt3}. 
As illustrated in Figure~\ref{fig:method_pipeline}, we randomly select  a scene and feed its one image into the BLIP to obtain its input caption, such as ``a blue and white vase."
Subsequently, we use GPT to rewrite the input caption to generate the target caption, such as ``a red and white vase."
For each image of a scene, we apply randomly minor perturbations by controlling the hyper-parameters $\theta$ in Eq.~\ref{eq:gen_edit}. For detailed parameter settings, please refer to the supplementary materials.

\noindent \textbf{Self-view robustness loss.}
We use a training approach similar to that in the generalizable NeRF models~\cite{wang2021ibrnet, liu2022neuray, johari2022geonerf}, which involves predicting the target view based on several source views. Unlike these methods, our training data incorporates minor perturbations.
In addition to reconstructing 3D scenes, our model's training objective also aims to remove these perturbations that cause inconsistencies.
To achieve this, we perform two independent predictions, labeled $\mathcal{A}$ and $\mathcal{B}$. 
For $\mathcal{A}$, we use source views $\{Src_a^1, Src_a^2,...\}$ to predict the target view $Tgt_a$. 
Additionally, for $\mathcal{B}$, we use different source views $\{Src_b^1, Src_b^2,...\}$ to predict the same target view $Tgt_b$. Then, we calculate the L2 loss to ensure consistency between the two predictions:
\\[-2pt]
\begin{equation}
\begin{split}
    \ell_{self} = \left \| Tgt_a - Tgt_b \right \|^2.
\end{split}
\label{eq:self_view}
\end{equation}
\\[-2pt]
\noindent \textbf{Neighboring view consistency loss}.
Empirically, we observe that there are often noticeable texture or color discontinuities between adjacent views when rendering along a smooth camera path.
To tackle this issue, we enforce a smooth transition between adjacent views. 
Specifically, we slightly perturb the pose corresponding to the target view and generate a  neighboring view $Tgt_{nbr}$ based on the perturbed pose. 
We then project the pixel points from the target view onto the neighboring view using the depth rendered by model $G$. We further minimize the following loss to reduce image discontinuities caused by changes in the viewing angle:
\\[-2pt]
\begin{equation}
\resizebox{.9\hsize}{!}{$
    \ell_{nbr} = \left \| M(Tgt_a - Tgt_{nbr}) \right \|^2 +  
                 \left \| M(Tgt_b - Tgt_{nbr}) \right \|^2,$}
\label{eq:neigh}
\end{equation}
\\[-2pt]
where $M$ refers to a mask that discards some pixels on the target view that are not visible to the neighboring view.

In addition, we note that the weights of some sampled points are uniformly distributed along the rays, leading to floating objects in the predicted results and inaccurate depth estimates, which may affect the accuracy of the target view to neighboring view projection.
Thus, we introduce an entropy loss~\cite{kim2022infonerf_enloss} for the weights of the sampled points:
\\[-2pt]
\begin{equation} 
\begin{split}
\resizebox{.9\hsize}{!}{$
    \ell_{en} =  -\sum T_i(1-\exp({-\sigma_i \delta_i}) \log{(T_i(1-\exp({-\sigma_i \delta_i}))},$}
\end{split}
\label{eq:loss_en}
\end{equation}
\\[-2pt]
where $T_i = \exp(- \sum_{j=1}^{i-1} \sigma_i \delta_i)$. $\sigma_i$ is the density of the sampled points and $\delta_i$ denotes the distance between adjacent sampled points.

\begin{figure*}[t]
        \centering
        \begin{center}
        \centerline{\includegraphics[width=0.93\textwidth]{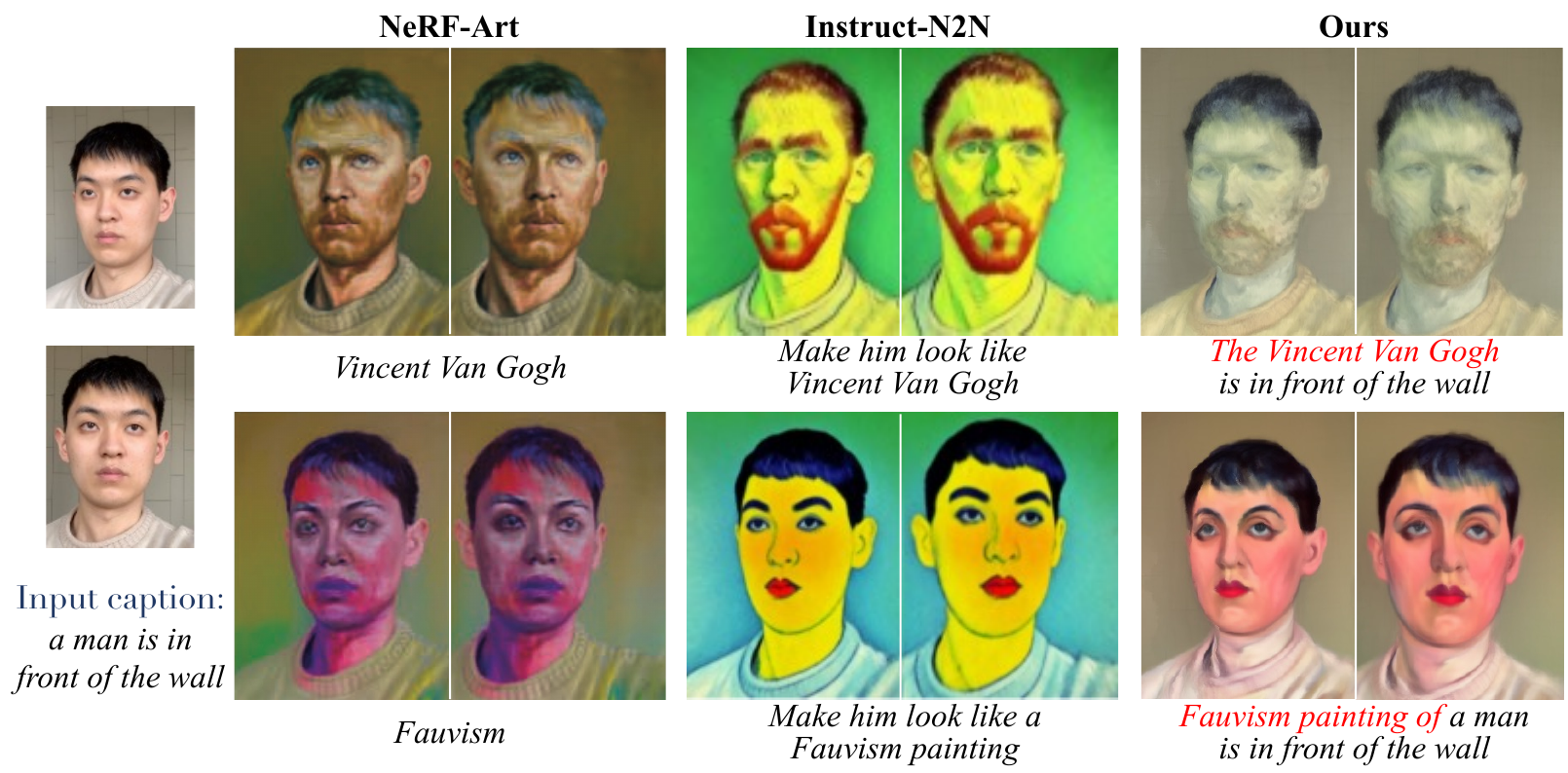}}
        \caption{\textbf{Comparison with other text-driven editing methods}. \modelname{} strikes a better balance between preserving image content and aligning with textual descriptions. More importantly, it is not necessary to retrain our model for different editing types. }
        \label{fig:text-methods-com}
        \end{center}
    \vspace{-5 ex}
\end{figure*}

\noindent \textbf{Total loss}. During training, the model cannot render a complete image in a single forward pass due to GPU memory limitations. 
Thus, the loss is computed on a patch level, and the total loss function employed in this work is:
\\[-2pt]
\begin{equation}
\begin{split}
    \ell = \ell_{rec} + \lambda_1 \ell_{self} + \lambda_2 \ell_{nbr} + \lambda_3 \ell_{en} + \lambda_4 \ell_{tv},
\end{split}
\label{eq:loss_all}
\end{equation}
\\[-4pt]
where $\ell_{rec}$ is the reconstruction loss obtained from Eq.~\ref{eq:argmin_real}, and $\ell_{tv}$ is the total variation regularization term~\cite{rudin1994tv_loss}.

\noindent \textbf{Content filter}.
During the inference stage, applying normal amplitude edits to 3D scene images may produce extremely poor editing results for certain views. These poor editing results cannot be fully remedied solely through the model $G$. Therefore, we designed a content filter first to remove these images.
After this step, only minor perturbations remain, which can be eliminated using the well-trained model $G$.
Combining evaluations of editing results from previous methods~\cite{wang2022nerfart,haque2023instruct}, we propose a content filter based on the following four tuples:
\\[-2pt]
\begin{equation}
\begin{aligned}
&1. \mathrm{SSIM}(I_i, \tilde I_i),~
  2. \mathrm{CLIP}(\tilde I_i,C_{tgt}),
  3. \mathrm{CLIP}(I_i, \tilde I_i),~\\
  &4.\mathrm{CLIP}(I_i, \tilde I_i)-\mathrm{CLIP}(C_{in}, C_{tgt}).
\end{aligned}
\label{eq:filter}
\end{equation}
\\[-2pt]
%
For all edited images, we compute the above four metrics and sort them individually. Then, we discard the top and bottom 10\% of outlier images for each metric, 
retaining only values close to the mean to reduce the differences among the edited images.
In our experiments, approximately 30\% of the images are ultimately filtered out. In cases where the discarded images exceed 50\%, we will employ the model $G$ to generate new images to mitigate data scarcity.
Filtered examples and numerical results for each metric can be found in the supplementary materials. 

\noindent \textbf{Network structure}. The generalizable NeRF model $G$ in Figure~\ref{fig:method_pipeline} is developed based on the IBRNet~\cite{wang2021ibrnet}. 
We extend it by incorporating multi-viewpoint aggregation, cross-view mappings, and integration of unedited image information to render consistent results. 
More details regarding the model $G$ are described in the supplementary materials.

\section{Experiments and Analysis}
The scenes used to train our model are from Google Scanned Objects~\cite{downs2022googlescan}, 
NeRF-Synthetic~\cite{mildenhall2020nerf},
Spaces~\cite{flynn2019space} and the IBRNet-collect~\cite{wang2021ibrnet}. 
The LLFF~\cite{mildenhall2019llff} and NeRF-Art~\cite{wang2022nerfart} datasets are used for evaluation. Also, all text prompts displayed in the experimental results are not used for training. Furthermore, we also verify DN2N on the OMMO~\cite{lu2023large-ommo} in the UAV's 360-degree view and the KITTI~\cite{geiger2013kitti} in the vehicle's view (The experimental results of these two data are shown in supplementary materials).
The default 2D-image editing model is Null-text~\cite{mokady2022null}.
We implement our method with PyTorch~\cite{paszke2019pytorch}, train the model on 8 NVIDIA V100 GPUs, and use one single V100 GPU for inference. For more implementation details, experimental results, and video demos, please see the supplementary materials.

\begin{figure*}[t]
        \centering
        \begin{center}
        \centerline{\includegraphics[width=0.98\textwidth]{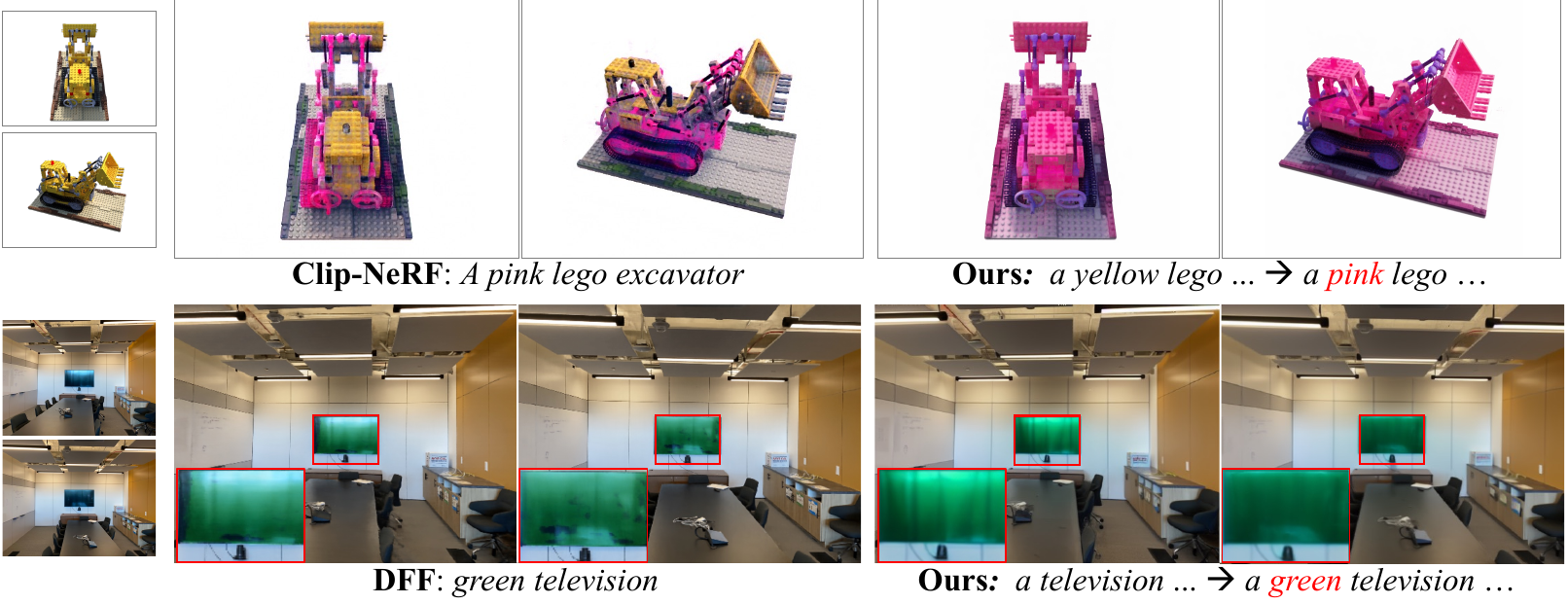}}
        \caption{\textbf{Comparison with other methods for editing appearance}. \modelname{} achieves higher accuracy in matching target captions 
        while effectively preserving information in non-edited areas.}
        \label{fig:appearance}
        \end{center}
    \vspace{-4 ex}
\end{figure*}

\begin{figure*}[t]
        \centering
        \begin{center}
        \centerline{\includegraphics[width=0.95\textwidth]{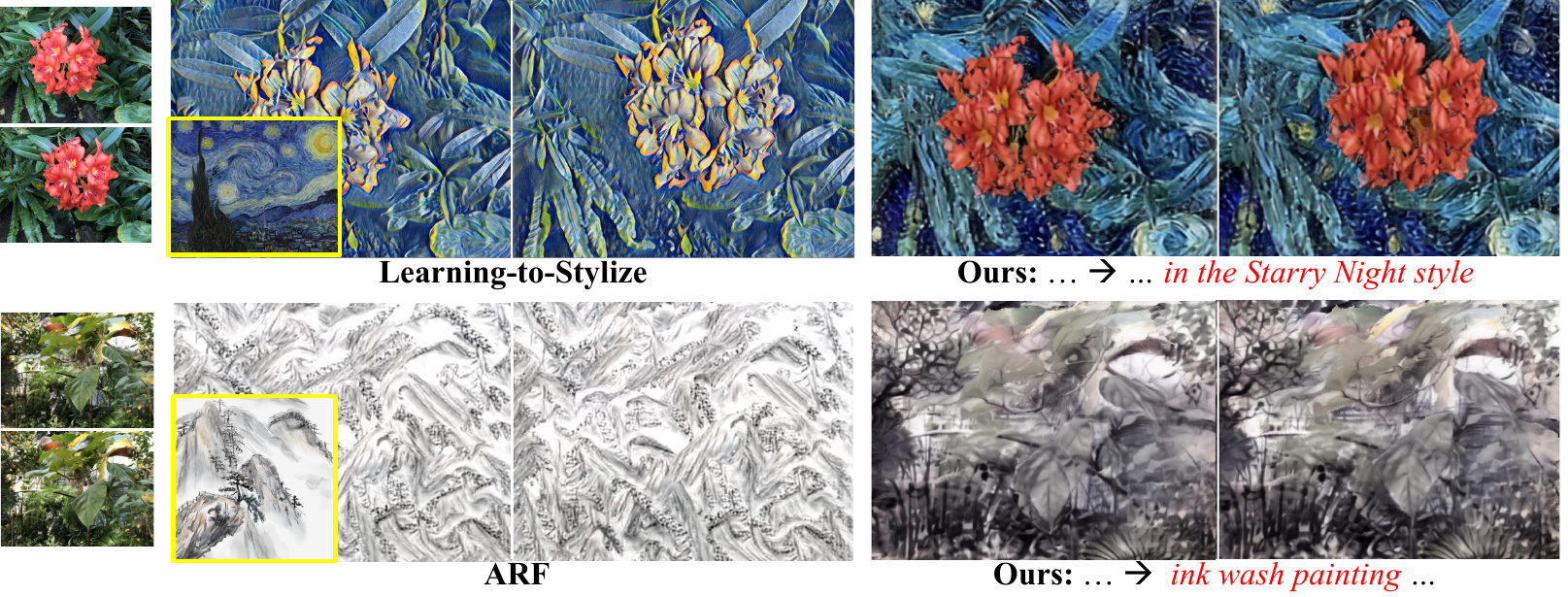}}
        \caption{\textbf{Comparison with other methods for editing scene styles}. \modelname{} can more accurately transfer colors and brush strokes while preserving more original image content.} 
        \label{fig:style-trans}
        \end{center}
    \vspace{-4 ex}
  \end{figure*}
  
\subsection{Qualitative results}
As illustrated in Figure~\ref{fig:shocking}, our approach demonstrates its ability to achieve various challenging editing types without retraining while maintaining 3D consistency and conforming to the text description.
To further assess the effectiveness of our approach, we conduct comparative studies against state-of-the-art methods.

\noindent \textbf{Text-driven 3D scene editing}.
We compare our approach to other text-driven editing methods in Figure~\ref{fig:text-methods-com}.
NeRF-Art utilizes multiple loss functions to retrain a model when given target words, with limitations in achieving complex and precise editing effects.
Instruct-N2N initially edits 2D images according to instruction and trains a model using these images, then generates new images to train the model recurrently until it produces images that comply with the instructions, which consumes a substantial amount of computational and storage resources.
In contrast, the scene editing results by \modelname{} are directly inferred after giving a text without any intervening training process. 
In addition, the editing results by \modelname{} are more realistic and retain more areas unrelated to the target caption, as shown in Figure~\ref{fig:text-methods-com}.

\begin{figure*}[htb]
\begin{minipage}{0.3\textwidth}
\captionof{table}{\textbf{Comparison of mean PSNR with other generalizable NeRF models on the LLFF dataset.} * indicates random minor perturbations to the scenes.}
  \label{tab:compare_nerf}
  \centering
  \setlength\tabcolsep{4pt}
  \resizebox{0.8\linewidth}{!}{
     \begin{tabular}{ccc}
    \toprule
          & LLFF  & $\mathrm{LLFF}^*$ \\
    \midrule
    PixelNeRF & 18.66 & 11.03 \\
    MVSNeRF & 21.18 & 16.74 \\
    IBRNet & 25.17 & 20.05 \\
    Neuray & \textbf{25.35} & 19.31 \\
    DN2N  & 23.81 & \textbf{22.42} \\
    \bottomrule
    \end{tabular}%
   }
\end{minipage}
\hspace{0.02\columnwidth}
\begin{minipage}{0.68\textwidth}
  \centering
  \includegraphics[width=1.\textwidth]{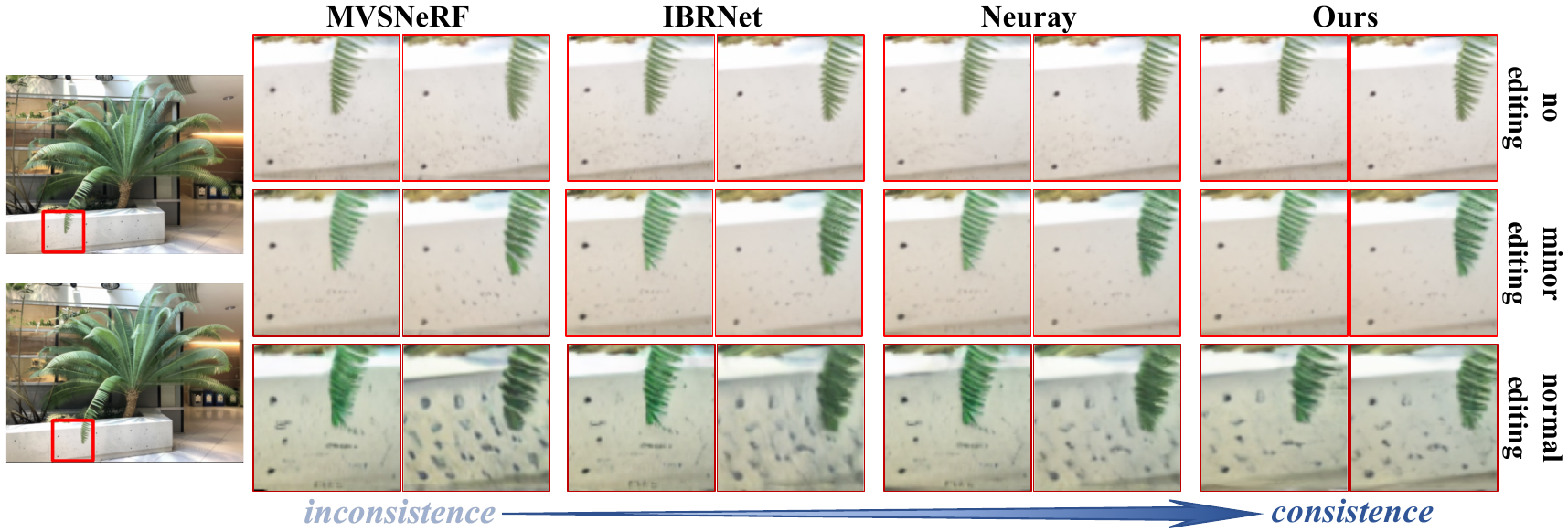}
          \caption{\textbf{Comparison with other generalizable NeRF models on the fern scene}. Our approach can maintain 3D consistency across novel views under different editing degrees.}
  \label{fig:general_model}
\end{minipage}

\vspace{-0 ex}
\end{figure*}

\noindent \textbf{Appearance editing}.
Two implicit steps are involved in editing the appearance of an object in a scene: determining the target area and editing the appearance of that area.
As illustrated in Figure~\ref{fig:appearance}, our method accurately locates the target area and applies appearance editing consistent with the target description without requiring training.
While Clip-NeRF uses CLIP Similarity to limit the novel view to the target words, it cannot accurately locate the target area.
DFF uses DINO or Lseg to inject label information into points in space, ensuring precise editing area localization.
However, its appearance editing requires manual operations, such as specifying the RGB value of the editing area. 
Furthermore, both methods require training separate models for each scene, making them less efficient than \modelname. 
  
\noindent \textbf{Style transfer}.
Figure~\ref{fig:style-trans} presents visual comparisons rendered by \modelname{} and other 3D scene style transfer techniques. 
As depicted, the Learning-to-Stylize method tends to imitate the color information of the reference image, but it often overlooks curved strokes. 
On the other hand, ARF excessively imitates curved strokes of the reference image, resulting in less pleasing visual effects and loss of information in the original scene. 
In contrast, our approach synthesizes the scenes by capturing both the color and stroke of the target style.
Furthermore, both Learning-to-Stylize and ARF necessitate selecting a reference image first and then training an editing model, making them less efficient and practical than the DN2N method.

\subsection{Quantitative results}
\textbf{Ability to resist perturbations}. 
Our method is specifically designed for scene editing, capable of maintaining 3D consistency under minor editing perturbations.
To demonstrate this, we compare our approach to commonly used generalization models in 8 scenes on the LLFF dataset.
Table~\ref{tab:compare_nerf} shows the results of two types of comparisons: one on unedited scenes (LLFF) and the other on scenes with minor editing perturbations ($\text{LLFF}^*$). 
It can be seen that our approach outperforms other models on scenes with minor perturbations resulting from 2D editing.
Figure~\ref{fig:general_model} also illustrates that our method can achieve superior 3D consistency in editing outcomes, which demonstrates our method is more suitable for the 3D scene editing task. 

\begin{table}[tb]
\centering
\caption{\textbf{Quantitative comparison of time and model size for editing the flower scene in LLFF dataset}. `TC' and `SC' stand for Time and Space Complexity, respectively.}
  \setlength\tabcolsep{3pt}
  \resizebox{0.98 \linewidth}{!}{

    \begin{tabular}{lcccccc}
    \toprule
          & \multicolumn{3}{c}{time (minute)} & \multirow{2}[4]{*}{TC} & \multirow{2}[4]{*}{Size(MB)} & \multirow{2}[4]{*}{SC} \\
\cmidrule{2-4}          & train & edit  & total &       &       &  \\
    \midrule
    ARF   & 21.7  & \textbf{3.4} & 25.1  & O(n)  & 558   & O(n) \\
    DFF   & 20.6  & 5.2   & 25.8  & O(n)  & 144   & O(n) \\
    Clip-NeRF & 524.4 & 349.7 & 874.1 & O(n)  & 29.4  & O(n) \\
    NeRF-Art & 1545.6 & 780.6 & 2326.2 & O(n)  & \textbf{18.3} & O(n) \\
    Instruct-N2N & 19.2  & 62.1  & 81.3  & O(n)  & 484   & O(n) \\
    Ours  & \textbf{0} & 22.3  & \textbf{22.3} & O(n)  & 103   & \textbf{O(1)} \\
    \bottomrule
    \end{tabular}%

}
\label{tab:time_space}
\end{table}

\noindent \textbf{Model efficiency}. 
Our generalizable model precludes retraining, which is more efficient and practical. A comparison of model efficiency has been incorporated, and the results, as depicted in Table~\ref{tab:time_space}, show substantial advantages of the proposed DN2N over state-of-the-art techniques in running time and model storage.
In terms of running time, our approach does not require training for new scenes or new types of edits, only necessitating inference time for editing, resulting in being more time-efficient than alternative techniques.
Regarding model storage, our model, which is applicable across all scenes and types of edits, entails a constant space complexity of $O(1)$, 
while other methods necessitate space complexity of $O(n)$.
When users intend to perform 100 different types of edits on 100 scenes, comparative methods necessitate the training and storage of 10,000 new models, whereas our method requires only one model.

\begin{figure*}[t]
        \centering
        \begin{center}
        \centerline{\includegraphics[width=0.98\textwidth]{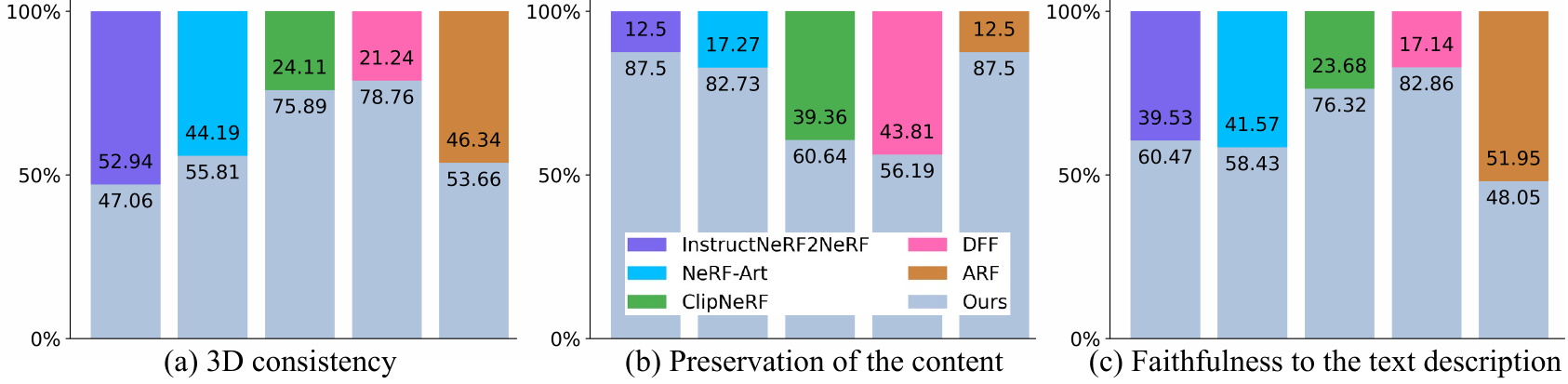}}
        \caption{\textbf{User study}. The proposed \modelname{} method performs well against other state-of-the-art approaches in terms of comprehensive performance across three evaluation criteria.}
        \label{fig:user_study}
        \end{center}
    \vspace{-5 ex}
\end{figure*}

\subsection{User study}
We perform a user study to further analyze the editing results of the proposed method against state-of-the-art approaches. 
We invited 50 participants to make a preference choice in editing results between DN2N and other methods, yielding 1700 votes in total for three evaluation metrics: 3D consistency, preservation of the original scene content, and faithfulness to the text description.
The results are depicted in Figure~\ref{fig:user_study}, which shows that DN2N is favored in terms of these evaluation metrics. The implementation details of the user study can be found in the supplementary materials.

\begin{figure}[t]
        \centering
        \begin{center}
    \centerline{\includegraphics[width=0.5\textwidth]{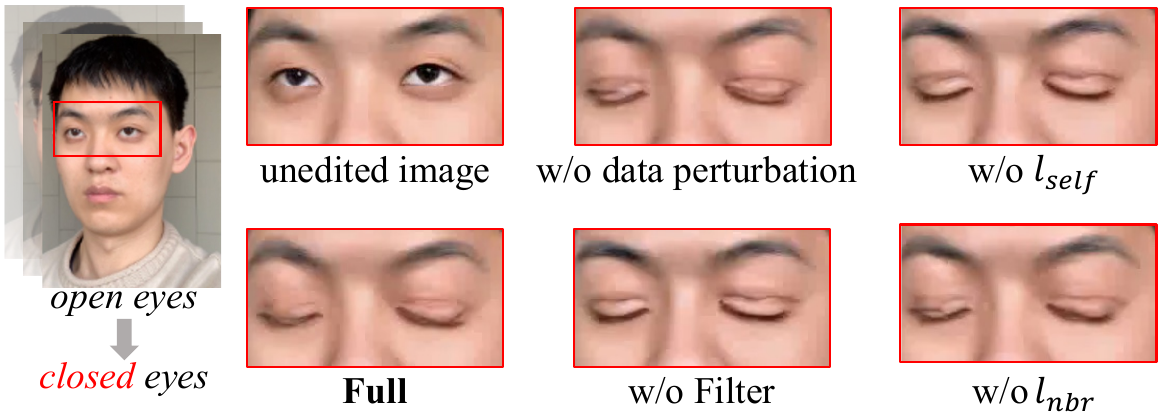}}
        \caption{\textbf{Ablation studies for key components in DN2N}. We recommend referring to the \textbf{\textit{video demo}} in the supplementary materials to observe the consistency differences.}
        \label{fig:ablation}
        \end{center}
    \vspace{-4 ex}
\end{figure}

\begin{table}[tb]
\centering
\captionof{table}{\textbf{PSNR results of ablation studies.} Experiments are conducted by applying four random minor perturbations to the scene editing in Figure~\ref{fig:ablation}. DP denotes data perturbation.}
  \label{tab:ablation}
  \centering
  \setlength\tabcolsep{5pt}
  \resizebox{0.8\linewidth}{!}{
    \begin{tabular}{c|ccccc}
    \toprule
          & w/o DP & w/o $l_{self}$ & w/o $l_{nbr}$  & Full \\
    \midrule
    exp1  & 17.24 & 19.89 & 21.05  & \textbf{21.76} \\
    exp2  & 18.98 & 21.5  & 21.92  & \textbf{23.11} \\
    exp3  & 18.73 & 21.16 & 22.39  & \textbf{24.21} \\
    exp4  & 18.97 & 20.53 & 22.19  & \textbf{22.27} \\
    \bottomrule
    \end{tabular}%
   }
    \vspace{-2 ex}
\end{table}

\subsection{Ablation studies}
We demonstrate the contribution of each component in our method in Figure~\ref{fig:ablation}. 
We find that the absence content filter results in significant image distortion.
This can be attributed to the fact that applying normal amplitude edits to 3D scene images may produce extremely poor editing results for a few views, making it difficult for the model to solve these inconsistencies without the content filter. 
Omitting the data perturbation process or removing the cross-view regularization terms during training would significantly affect the model performance on edited results. 
Furthermore, we directly evaluate DN2N's ability to remove perturbations by applying several random disturbances to the scene depicted in Figure~\ref{fig:ablation}. The PSNR results are presented in Table~\ref{tab:ablation}.
It is clear that training the generalization model with perturbed data is crucial and enforcing cross-view consistency can also enhance the model's overall performance.

\subsection{Limitations} \label{subsec:limitation}
\begin{figure}[t]
        \centering
        \begin{center}
    \centerline{\includegraphics[width=0.47\textwidth]{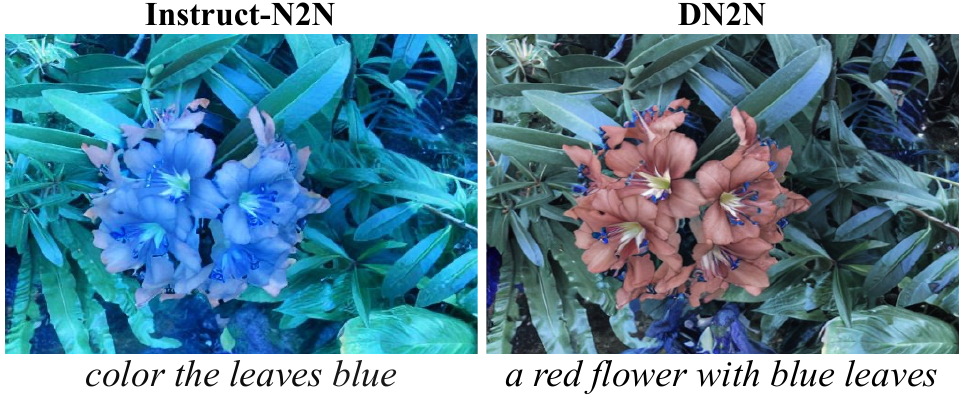}}
        \caption{\textbf{Failed editing cases}. The editing result is limited by the 2D multimodal model.}
        \label{fig:attention}
        \end{center}
    \vspace{-4 ex}
\end{figure}

When utilizing 2D multimodal models for 3D scene editing, a common limitation arises wherein the 3D editing outcomes heavily depend on 2D methods, making it challenging to achieve editing beyond the capabilities of the 2D model. For instance, CLIP-NeRF~\cite{wang2022clipnerf} is limited by the CLIP model~\cite{radford2021clip}, while Instruct-N2N~\cite{haque2023instruct} is influenced by the results of InstructPix2Pix~\cite{brooks2022instructpix2pix}. Similarly, our approach is also constrained by Null-text~\cite{mokady2022null}.
As shown in Figure~\ref{fig:attention}, 2D editing models may not always achieve reliable editing results, leading to failures in editing 3D scenes. 

\section{Conclusion}
In this work, we propose a text-driven method for editing 3D scenes that exhibits strong generalization capabilities and enables realistic novel view editing without additional training for each modification task.
Our approach leverages existing 2D editing models to perform initial editing of 3D scene data based on textual prompts. 
We then filter out poorly edited images, treating the remaining inconsistency as perturbations on top of consistently edited results. 
To eliminate this perturbation, we provide several approaches to train our model, including creating training data and strengthening cross-view robustness.
The experimental results demonstrate the effectiveness of our approach, which offers significant advantages over other methods. 
Specifically, our method provides greater convenience for editing, supports multiple editing capabilities, and eliminates the need for training on new scenes, thus significantly reducing the editing time and model storage requirements.

{
    \small
    \bibliographystyle{unsrt}
    \bibliography{main}
}


\clearpage
\setcounter{page}{1}
\maketitlesupplementary

\newcommand{\eps}{\varepsilon}
\newcommand{\norm}[1]{\left\Vert #1 \right\Vert_2}
\newcommand{\textemb}{\mathcal{C}}

\section{Background of neural radiance fields and diffusion models}

\textbf{Neural radiance fields}. NeRF utilizes an implicit function to represent scenes, which maps the spatial point $\mathbf{x}=(x, y, z)$ and view direction $\mathbf{d}=(\theta, \phi)$ to the density $\sigma$ and color $\mathbf{c}$. 
Typically, the implicit function is represented by an MLP network, denoted as 
 $F_\Theta : (\mathbf{x}, \mathbf{d}) \longrightarrow  (\sigma, \mathbf{c})$, where $\Theta$ represents the weights of the network. 
For a ray $\mathbf{r}$ originating at $\mathbf{o}$ with direction $\mathbf{d}$, the RGB value $\hat{\mathbf{C}}(\mathbf{r})$ of the ray is estimated through numerical quadrature of the color $\mathbf{c}$ and density $\sigma$ of the $N$ spatial points.
\\[-3pt]
\begin{equation}  
    \hat{\mathbf{C}}(\mathbf{r}) = \sum_{i}^{N}T_i(1-\exp(-\sigma_i \delta_i))\mathbf{c}_i,
    \label{neural_rendering_equation}
\end{equation}
\\[-3pt]
where $\delta_i$ denotes the distance between adjacent sample points, and $T_i = \exp(- \sum_{j=1}^{j=i-1} \sigma_j \delta_j)$.

\noindent \textbf{Text-guided diffusion models.}
Text-guided diffusion models aim to generate an output image $z_0$ from a random noise vector $z_t$ under a textual condition $\mathcal{P}$. To achieve sequential denoising, the model $\eps_\theta$ is trained to predict artificial noise, minimizing the objective:
\\[-3pt]
\begin{equation}
\min_\theta E_{z_0,\eps\sim N(0,I),t\sim \text{Uniform}(1,T)} \norm{\eps - \eps_\theta(z_t,t,\textemb)}^2,
\end{equation}
\\[-3pt]
where $\textemb$ denotes the embedding of the text condition, and $z_t$ is a noised sample according to the timestamp $t$. 
When inference, given a vector $z_T$, its noise is gradually removed by sequential prediction using a trained network for $T$ steps~\cite{song2020denoising}.
%
Amplifying the effect induced by the conditioned text is a significant challenge in a text-guided generation. To address this issue, Ho et al.~\cite{ho2022classifier} propose a guidance technique that eliminates the need for a classifier in unconditional prediction and extends it to conditioned prediction scenarios.
With the introduced concept of a null text embedding, denoted as $\varnothing$,
and a guidance scale parameter $w$, then the guidance prediction is given by:
\\[-3pt]
\begin{equation}
\tilde{\eps}_\theta(z_t,t,\textemb,\varnothing) = w \cdot \eps_\theta(z_t,t,\textemb) + (1-w) \cdot \eps_\theta(z_t,t,\varnothing) .
\end{equation}
\\[-3pt]
In our experiments, we primarily control the degree of image editing by adjusting the parameters $w$ and $T$.

\begin{figure*}[!ht]
        \centering
        \begin{center}
        \centerline{\includegraphics[width=0.92\textwidth]{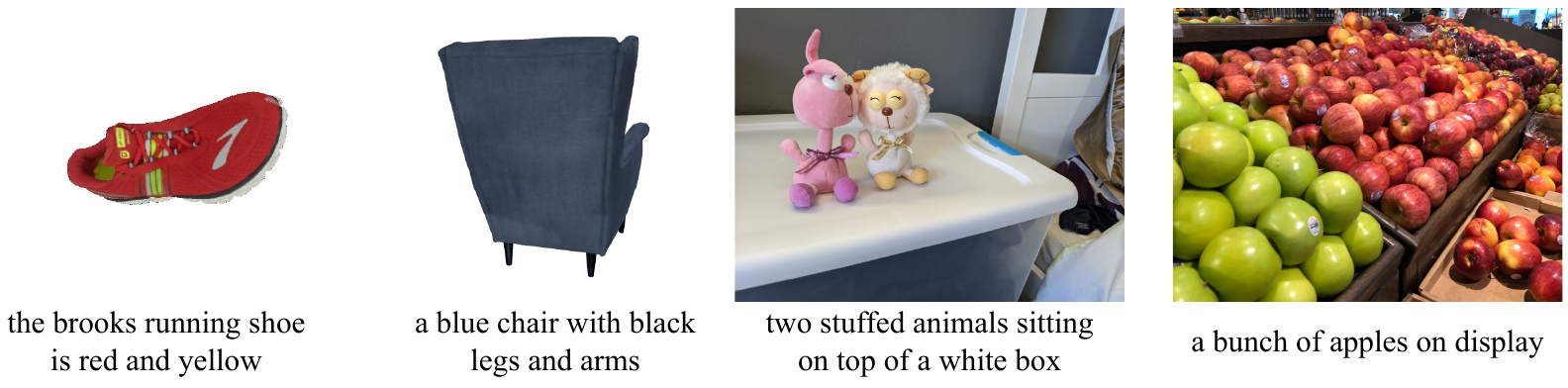}}
        \caption{\textbf{Generate input captions by BLIP}. We randomly select an image in a 3D scene to generate its input caption by BLIP~\cite{li2023blip2}.}
        \label{fig:blip_caption}
        \end{center}
    \vspace{-3 ex}
  \end{figure*}

\begin{table*}[!htbp]
  \centering
  \caption{\textbf{Generate target captions by GPT}. Given the input caption \textcolor{blue}{S} for a 3D scene, we utilize GPT~\cite{brown2020gpt3} to generate a target caption \textcolor{red}{O}. For instance, if \textcolor{blue}{S} is \textquotesingle \emph{yellow roses in the garden} \textquotesingle, the target captions can be 
  \textquotesingle \emph{Leonardo da Vinci painting of yellow roses in the garden} \textquotesingle, 
  \textquotesingle \emph{yellow roses in the garden in the Rococo style}\textquotesingle, 
  \textquotesingle \emph{pink roses in the garden} \textquotesingle, to name a few.}
  \label{tab:mytable}
  \begin{adjustbox}{max width=1.\textwidth}
    \begin{tabular}{C{1.3cm}C{2.7cm}C{2.7cm}C{2.5cm}C{4.4cm}}
    \toprule
    Prompts & List 100 famous painters & List 50 famous painting schools & List 100 famous paintings & Replace, add or delete partial words in the following sentences: \textcolor{blue}{S1}, \textcolor{blue}{S2}, ... \\
    \midrule
    GPT \newline outputs \newline (\textcolor{red}{O})
    &
    Leonardo da Vinci  \newline
    Vincent van Gogh \newline
    Pablo Picasso \newline
    \enspace . \newline
    \ . \newline
    Henri Matisse \newline
    Eva Hesse \newline
    Carl Andre
    &
    Baroque \newline
    Realism \newline
    Impressionism \newline
    . \newline
    . \newline
    Tonalism \newline
    Ashcan School \newline
    Rococo
    & 
    Mona Lisa \newline
    The Last Supper \newline
    The Scream \newline
    . \newline
    . \newline
    The Fifer \newline
    The Kiss \newline
    The Hay Wagon
    & 
    pink roses in the garden \newline
    red roses in the garden \newline
    white roses in the garden \newline
    . \newline
    . \newline
    a green couch with gold trim \newline
    a green  chair with silver trim \newline
    a green chair with no trim
     \\
    \hline
    Target \newline captions & 
    \textcolor{red}{O} painting of \textcolor{blue}{S} \newline
    or \newline
    \textcolor{blue}{S} in the \textcolor{red}{O} style 
    &
    \textcolor{red}{O} painting of \textcolor{blue}{S} \newline
    or \newline
    \textcolor{blue}{S} in the \textcolor{red}{O} style 
    &
    \textcolor{blue}{S} in the \textcolor{red}{O} style 
    & \textcolor{red}{O} \\
    \bottomrule
    \end{tabular}

    \end{adjustbox}
    \label{tab:gpt_caption}
    \vspace{2 ex}
\end{table*}

\begin{figure*}[!htbp]
        \centering
        \begin{center}
        \centerline{\includegraphics[width=0.75\linewidth]{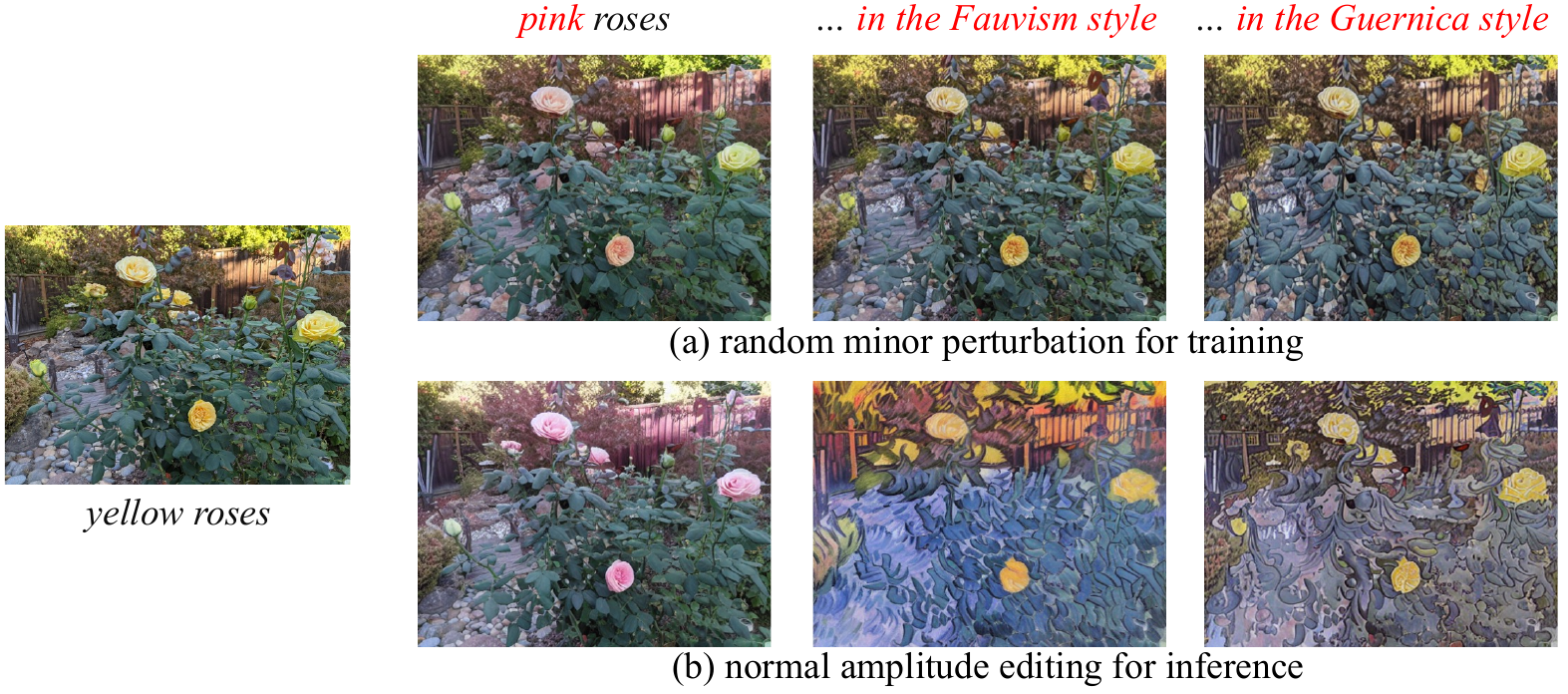}}
        \caption{\textbf{Visualization of random minor perturbations and normal amplitude editing}. The proposed model learns to remove the added perturbations on (a). Upon completion of the learning process, the model generates consistent editing results by eliminating inconsistencies between images subjected to normal amplitude editing on (b).}
        \label{fig:vis_pertur}
        \end{center}
    \vspace{-3 ex}
\end{figure*}

\section{Network architecture}
Initially, we utilize a UNet-like network with the ResNet34~\cite{he2016resnet} backbone to extract features from both the unedited and edited source views. These feature maps are then concatenated and inputted into subsequent CNN networks to derive the final feature maps. When estimating the color $\mathbf{c}$ and density $\sigma$ of sampled points along the rays, we primarily employ the MLP networks to integrate information from the edited images, feature maps, and viewing directions. For density prediction, we adopt a design of transformer networks inspired by IBRNet~\cite{wang2021ibrnet}. Regarding the color estimation, we incorporate additional pixel difference information among source views to capture the extent of editing across different perspectives, aiding the network in improved learning. The detailed network architecture and data propagation process are depicted in Figure~\ref{fig:network}.

\section{Additional implementation details} \label{sec:implementation}
\subsection{Generating training data pairs} \label{sec:generate}
\textbf{Generate input and target captions}. During the training phase, we utilized a total of 1246 scenes. For each scene, we randomly select one image to generate an input caption using the BLIP model~\citep{li2023blip2} with 2.7 billion parameters~\footnote{https://huggingface.co/Salesforce/blip2-opt-2.7b}. A subset of the generated captions is shown in Figure~\ref{fig:blip_caption}.
Then we utilize the GPT model~\citep{brown2020gpt3} to generate target captions.
As shown in Table~\ref{tab:gpt_caption}, the four instruction prompts for GPT are:

(1). List 100 famous painters.

(2). List 50 famous painting schools.

(3). List 100 famous paintings.

(4). Replace, add, or delete partial words in the following sentences: X. (X is the input caption from BLIP.)

During training, we select one of the above (1)-(4) following a 2:2:2:4 ratio to generate the target caption. For (1), a chosen painter like ``Van Gogh" could transform the caption ``a red flower" into ``Van Gogh painting of a red flower". Similar procedures apply to (2) and (3). For (4), GPT might change ``a red flower" to ``a red apple". Thus, each scene incorporates 405 target captions.

\noindent \textbf{Minor perturbations for training}.
We generate the training data by applying minor random perturbations to the 3D scene using the Null-text~\footnote{https://null-text-inversion.github.io}. The relevant parameters are set as follows: 100 to 300 for the iteration number $T$ and 0.5 to 3.5 for the text guidance scale $w$.

\noindent \textbf{Normal perturbations for inference}.
For normal-scale edits, we follow the recommended settings from the Null-text paper, with $w$=7.5 and $T$=500. 

The effects of applying random minor perturbations and normal editing to the images are depicted in Figure~\ref{fig:vis_pertur}.

\begin{figure*}[t]
        \centering
        \begin{center}
        \centerline{\includegraphics[width=0.8\linewidth]{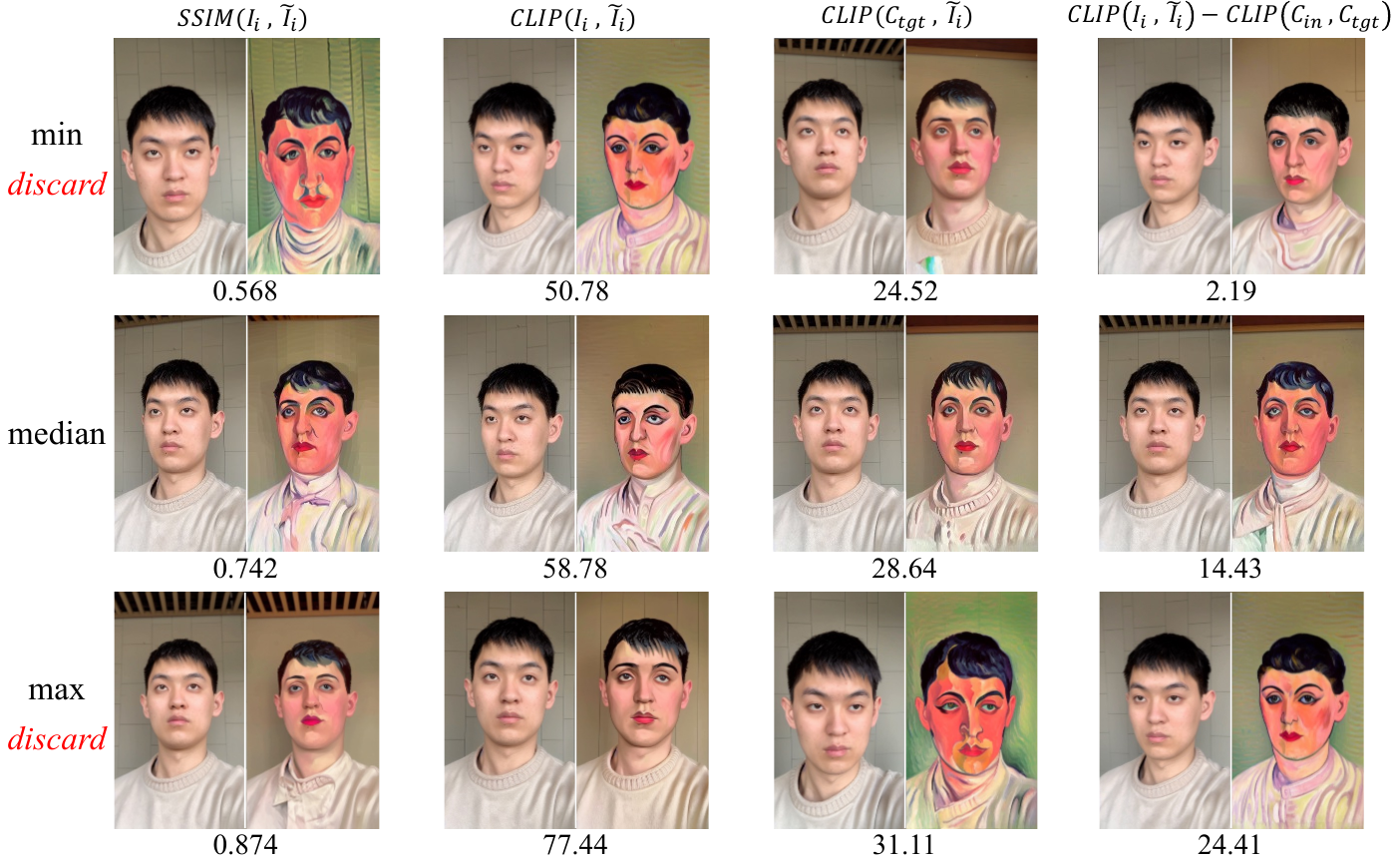}}
        \caption{\textbf{Visualize the editing results of the 2D editing model, as well as the maximum, minimum, and median values for each metric of the content filter}. The images that correspond to the minimum and maximum values often exhibit either low or excessive editing degrees. By discarding them, we can enhance the 3D consistency among the remaining images.}
        \label{fig:filter_example}
        \end{center}
    \vspace{-3 ex}
\end{figure*}

\subsection{Details of training and inference}
\textbf{Training phase}. 
Our method is implemented using the PyTorch framework \cite{paszke2019pytorch}. We employ the Adam optimizer \cite{kingma2014adam} with initial learning rates of 1e-4 for the CNN and 5e-4 for the MLP. The training process runs for 300K steps with a batch size of 500 rays. The initial values for the loss weight in Eq.~8 in the main text
of $l_{self}$, $l_{nbr}$, $l_{en}$, and $l_{tv}$, are set as $\lambda_1$=1e-3, $\lambda_2$=1e-3, $\lambda_3$=1e-3, and $\lambda_4$=2e-3, respectively. The calculation of $l_{nbr}$ is only performed after the iteration count exceeds 10K. During the training phase, we randomly select a variable number of source views ranging from 6 to 15, while using 15 source views during inference. The number of sampled points on a ray is set to 64.

\noindent \textbf{Inference phase}. 
After completing the training process, given a scene and its corresponding textual description, we apply normal-level editing to the images of the scene. Following that, we employ a content filter to select the edited results, removing the lowest and highest 10\% of values for each of the four metrics defined in Eq.~9 in the main text.

\noindent \textbf{Implementation details of the content filter}.
The main idea behind the content filter is to maintain the degree of editing consistent across various perspectives. To achieve this, we consider the following two situations:

(1). A smaller degree of editing implies fewer changes in the \textit{edited image} relative to the \textit{original image} and \textit{input caption}, while a larger discrepancy with the \textit{target text}.

(2). Conversely, a higher degree of editing denotes more significant changes in the \textit{edited image} compared to the \textit{original}, bringing it closer to the \textit{target text}.

Therefore, the evaluation metrics used in the content filter can be measured through the relationship between the original image, edited image, original text description, and target text description. 
Specifically, given an original image $I_i$ and its caption $C_{in}$,  as well as its corresponding edited image $\tilde I_i$ and its caption $C_{tgt}$, we calculate the following four measurements based on SSIM similarity and CLIP similarity during the filtering process:

(a). $\mathrm{SSIM}(I_i, \tilde I_i)$

(b). $\mathrm{CLIP}(I_i, \tilde I_i)$

(c). $\mathrm{CLIP}(\tilde I_i,C_{tgt})$

(d). $\mathrm{CLIP}(I_i, \tilde I_i) - \mathrm{CLIP}(C_{in}-C_{tgt})$

Here, the metrics (a) and (b) assess the similarity between the image before and after editing. The metric (c) gauges the similarity between the edited image and the target caption, while the metric (d) evaluates the relative offset between the text and the image. These indicators assess the editing results on different dimensions. The role of our content filter is to eliminate extreme values (top 10\% and bottom 10\%) from these metrics, ensuring metric values between the remaining images are not significantly different.

\begin{table*}[t]
\centering
\caption{\textbf{Quantitative comparison of CLIP Directional Score (CDS) and CLIP Consistency Score (CCS)}. $\text{CDS}^*$ indicates using a different caption from CDS. $\text{CCS}^*$ means our result at the minimum editing degree. \textit{The displayed values are multiplied by 100}.}
\setlength{\tabcolsep}{4.5pt}
  \resizebox{0.6\linewidth}{!}{
    \begin{tabular}{lcccc|cccc}
    \toprule
          & \multicolumn{4}{c|}{Van Gogh} & \multicolumn{4}{c}{Fauvism} \\
    \midrule
          & CDS   & CDS*  & CCS   & CCS*  & CDS   & CDS*  & CCS   & CCS* \\
    NeRF-Art & \textbf{13.07} & 10.21 & 2.06  & 2.06  & \textbf{17.38} & 12.74 & 1.38  & 1.38 \\
    Instruct-N2N & 11.21 & 10.39 & \textbf{9.24} & \textbf{9.24} & 16.62 & 13.18 & \textbf{4.11} & 4.11 \\
    Ours  & 8.83  & \textbf{11.76} & 4.89  & \textbf{100} & 14.06 & \textbf{14.93} & 2.29  & \textbf{100} \\
    \bottomrule
    \end{tabular}%

  }
\label{tab:cds_ccs_metric}
\end{table*}

In Figure~\ref{fig:filter_example}, we provide the maximum, minimum, and median values for each of the aforementioned metrics, along with their corresponding original and edited images. Figure~\ref{fig:filter_example} shows that the editing results with maximal and minimal metric values exhibit larger image discrepancies. These will pose greater challenges for subsequent 3D consistency. Our content filter, by eliminating these extreme editing results, facilitates superior 3D editing outcomes.

\begin{figure*}[hbt]
        \centering
        \begin{center}
       \centerline{\includegraphics[width=0.75\linewidth]{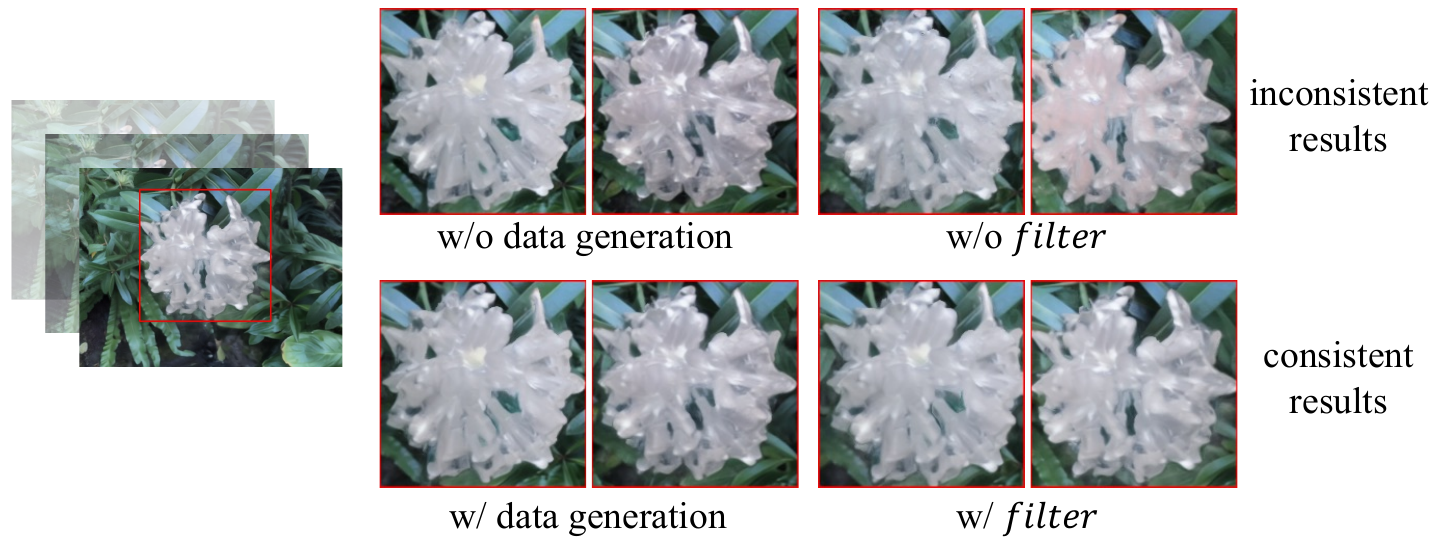}}
        \caption{\textbf{Ablation study on data generation and content filter}. Without incorporating data augmentation involving minor perturbations and the content filter component, the rendered novel views may exhibit noticeable inconsistencies in terms of glossiness and color.}
        \label{fig:ablation_dg_filter}
        \end{center}
    \vspace{-4 ex}
\end{figure*}

\begin{figure*}[hbt]
        \centering
        \begin{center}
       \centerline{\includegraphics[width=0.75\linewidth]{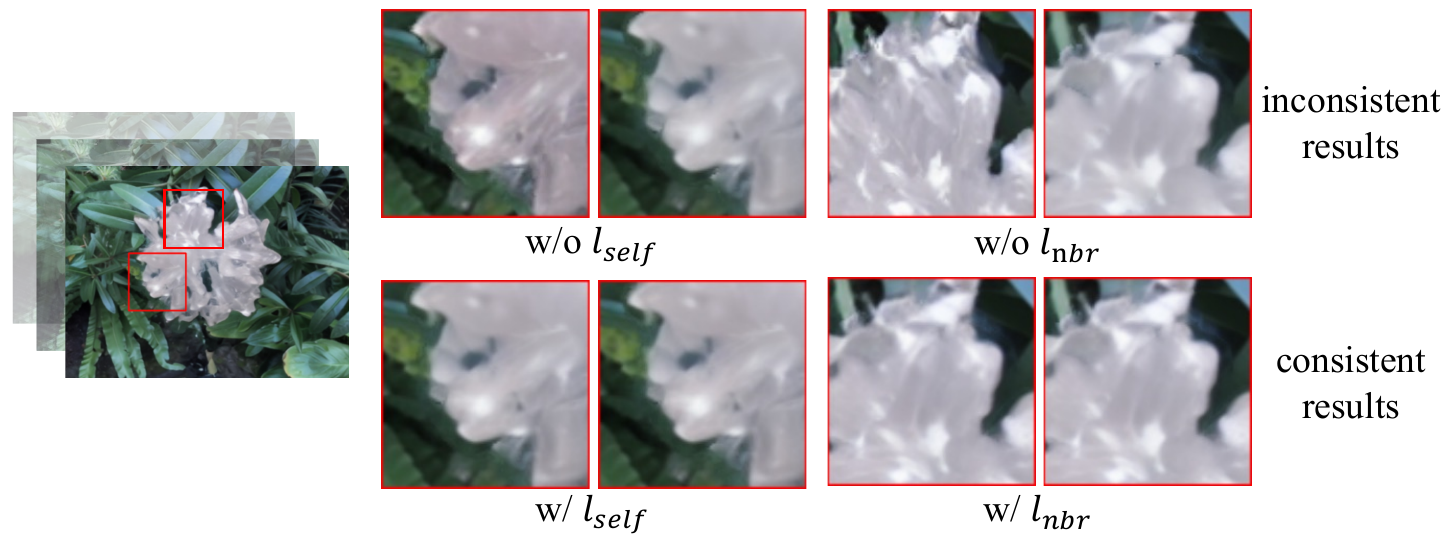}}
        \caption{\textbf{Ablation study on $l_{self}$ and $l_{nbr}$ regularization terms}. Without the $l_{self}$, the 3D consistency of the rendering results under different source views is poor. And without the $l_{nbr}$, inconsistency exists among different neighboring views.}
        \label{fig:ablation_self_nbr}
        \end{center}
    \vspace{-4 ex}
\end{figure*}

\section{Additional quantitative comparison}

Although the evaluation of the editing results is subjective, we follow the descriptions of CLIP Directional score (CDS) and CLIP Consistency score (CCS) as outlined by Instruct-N2N~\citep{haque2023instruct} to quantitatively evaluate the editing results shown in Figure 3 of the main text. The quantitative results are presented in Table~\ref{tab:cds_ccs_metric}. It is clear that, under different settings, there are large variances in these metrics.

The CDS measures how much the change in text captions agrees with the change in images. When using CDS, different descriptions yield varying results, as shown by CDS and CDS* in Table~\ref{tab:cds_ccs_metric}. This is mainly because the prompts used in training vary across different methods. 
For example, NeRF-Art just uses target words like “Van Gogh”, and incorporates the CDS into the loss function (as in Equation 4 of the NeRF-Art paper), while Instruct-N2N employs instructional prompts such as “Make him look like Vincent Van Gogh”. Our method, on the other hand, provides a target description like “Vincent Van Gogh is in front of the wall.” 
Thus, providing an equitable text description for comparing CDS presents a challenge. With the description “Portrait of Van Gogh”, NeRF-Art has a higher CDS. When “Vincent Van Gogh is in front of the wall” is provided, our method performs the best. Therefore, CDS may not fully evaluate image editing performance effectively.

The CCS measures the cosine similarity of the CLIP embedding of each pair of adjacent frames in a novel camera path. This metric heavily depends on the degree of editing. In extreme cases where the edited result is identical to the unedited one, the CCS has a max value (close to 1). However, under such conditions, the editing effect is not achieved. Therefore, the CCS also has certain limitations.

Due to these considerations, both DN2N and Instruct-N2N mainly compare with other methods for visual effects, rather than for CDS and CCS metrics. Effective quantitative comparison for editing results remains challenging, so existing methods still opt for subjective evaluations, for instance, by conducting user studies to aggregate subjective evaluation results from multiple individuals in order to reflect the quality of the editing results.

\begin{figure}[hbt]
        \centering
        \begin{center}
        \centerline{\includegraphics[width=0.5\textwidth]{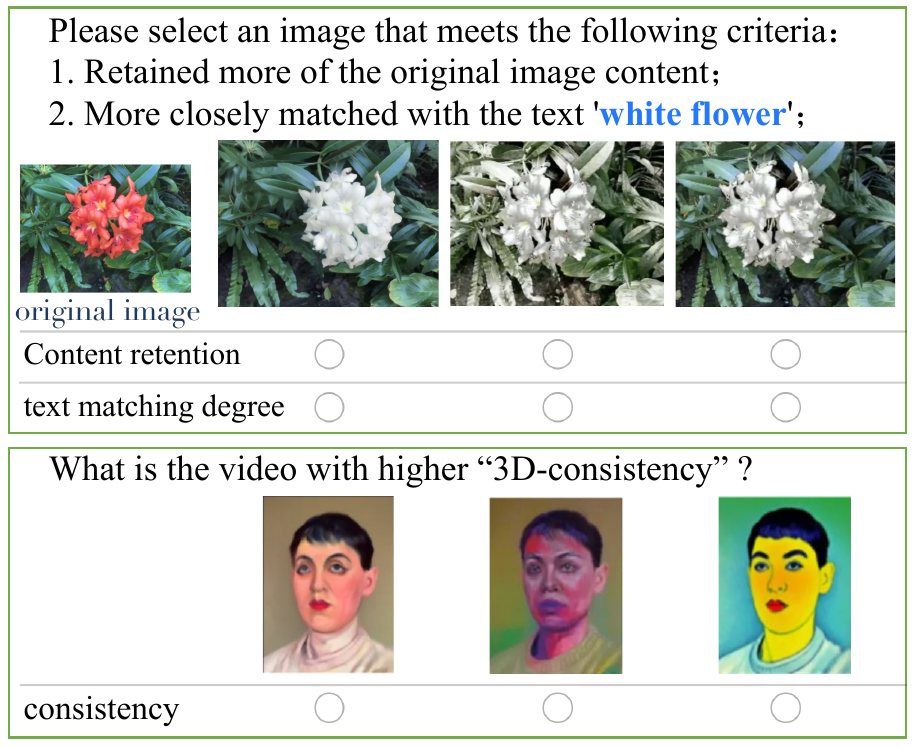}}
        \caption{\textbf{Example of questionnaire for user study}.}
        \label{fig:user_study_screen}
        \end{center}
    \vspace{-4 ex}
  \end{figure}

\section{Additional ablation studies} \label{sec:more_ablation}
In this section, we present additional ablation experiments related to the training data generation, content filter, self-view ($l_{self}$), and neighboring-view ($l_{nbr}$) regularization terms.
Specifically, we edit the flower scene in the LLFF dataset to a transparent ice sculpture flower. The ablation results are shown in Figure~\ref{fig:ablation_dg_filter} and Figure~\ref{fig:ablation_self_nbr}

Without perturbed data generation, we observed significant color and gloss discontinuities between different novel views in the editing results as shown in Figure~\ref{fig:ablation_dg_filter}. This is mainly due to the model not learning how to remove inconsistent perturbations between images.
In the inference stage, the normal editing may cause substantial consistency disruption in 2D editing results, this is mitigated by the content filter. Removing this filtering process leads to noticeable inconsistencies in the results as depicted in Figure~\ref{fig:ablation_dg_filter}.
The situation in Figure~\ref{fig:ablation_self_nbr} is similar, noticeable inconsistencies in color or brightness can be seen when removing the $l_{self}$ and $l_{nbr}$ regularization terms.

\section{User study details}
We invited a total of 50 subjects (31 males and 19 females) to participate in the user study.
Each participant is presented with videos or frames generated by various methods. They are asked to select their preferred option based on three distinct criteria.
%
An example of the questionnaire is available in Figure~\ref{fig:user_study_screen}.
Typical questions in the questionnaire are:

(1). What is the video with higher consistency? 

(2). Please select an image that retains more of the original image content.

(3). Please select an image that more closely matches the text description.

We use 23 videos to evaluate 3D consistency, and 24 images to measure content preservation and faithfulness to the text description.
After completing the survey, we tally user support rates for our method compared to other methods, as shown in Figure 7 in the main text.

\begin{figure*}[htbp]
        \centering
        \begin{center}
       \centerline{\includegraphics[width=0.92\linewidth]{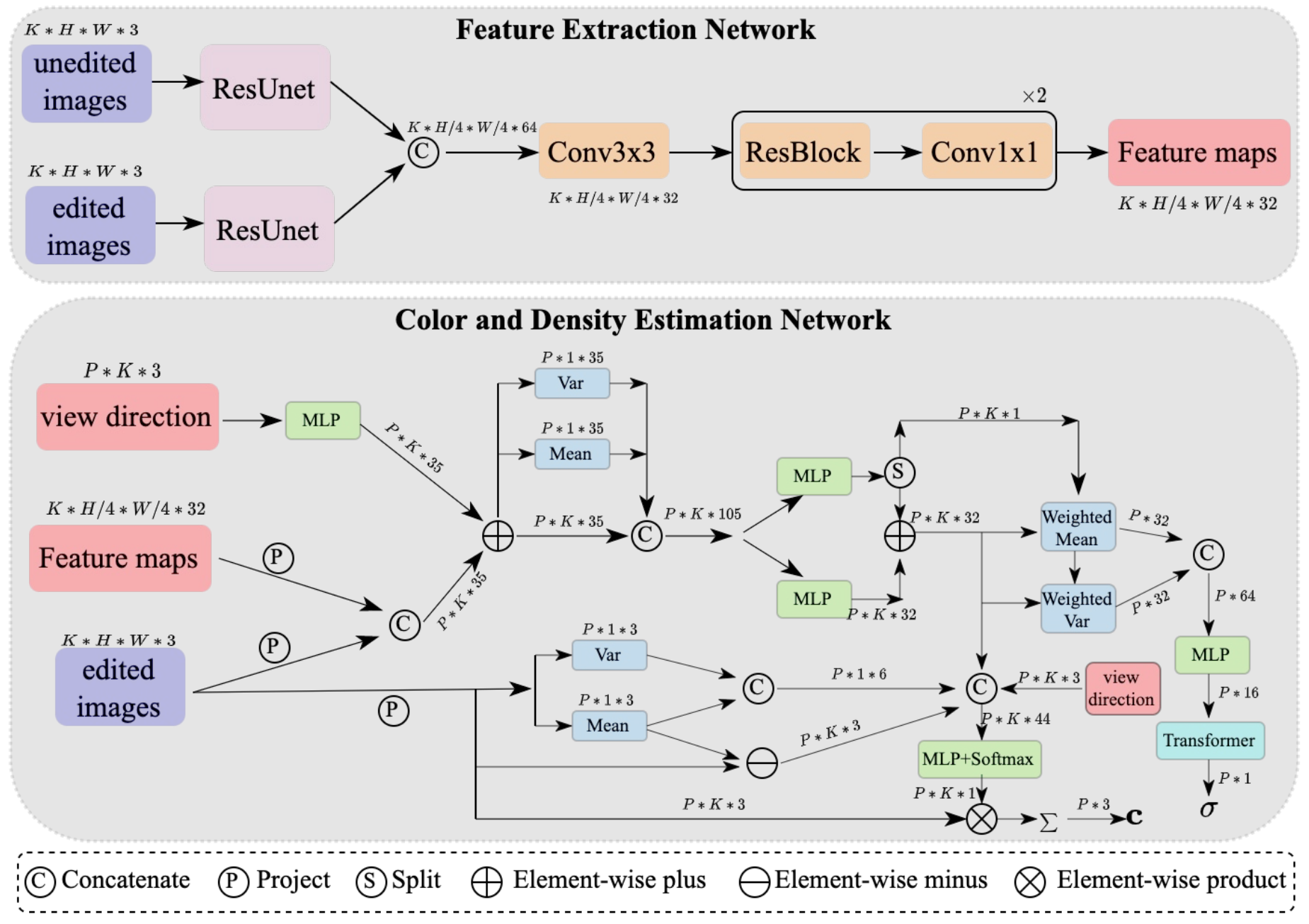}}
        \caption{\textbf{Network architecture}. $P$ is the number of sample points on a ray, and $K$ is the number of source views.}
        \label{fig:network}
        \end{center}
\end{figure*}

\begin{figure*}[t]
        \centering
        \begin{center}
        \centerline{\includegraphics[width=0.9\textwidth]{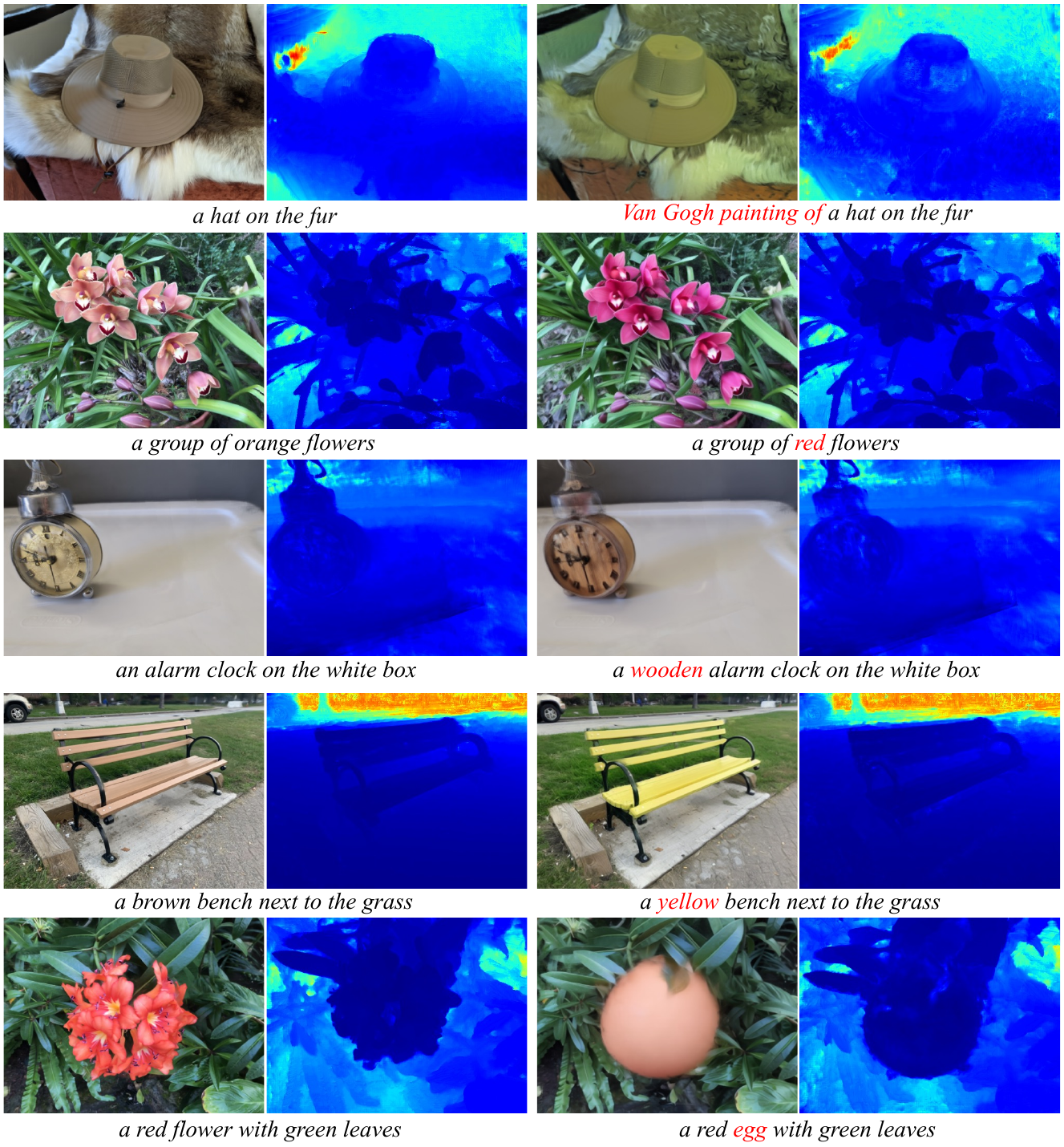}}
        \caption{\textbf{Visualization of geometric information}.}
        \label{fig:vis_depth}
        \end{center}
    \vspace{-4 ex}
  \end{figure*}

\section{Visualization of geometric information}
We visualize the depth maps before and after scene editing, as shown in Figure~\ref{fig:vis_depth}. This is divided into two cases: first, when only appearance is edited, the depth maps before and after exhibit negligible differences. Second, for edits altering the geometry of objects, the depth maps correspondingly exhibit changes as well, validating the 3D awareness of our editing approach.

\section{Additional editing results} \label{sec:more_res}
\begin{figure*}[htbp]
        \centering
        \begin{center}
        \centerline{\includegraphics[width=1.\linewidth]{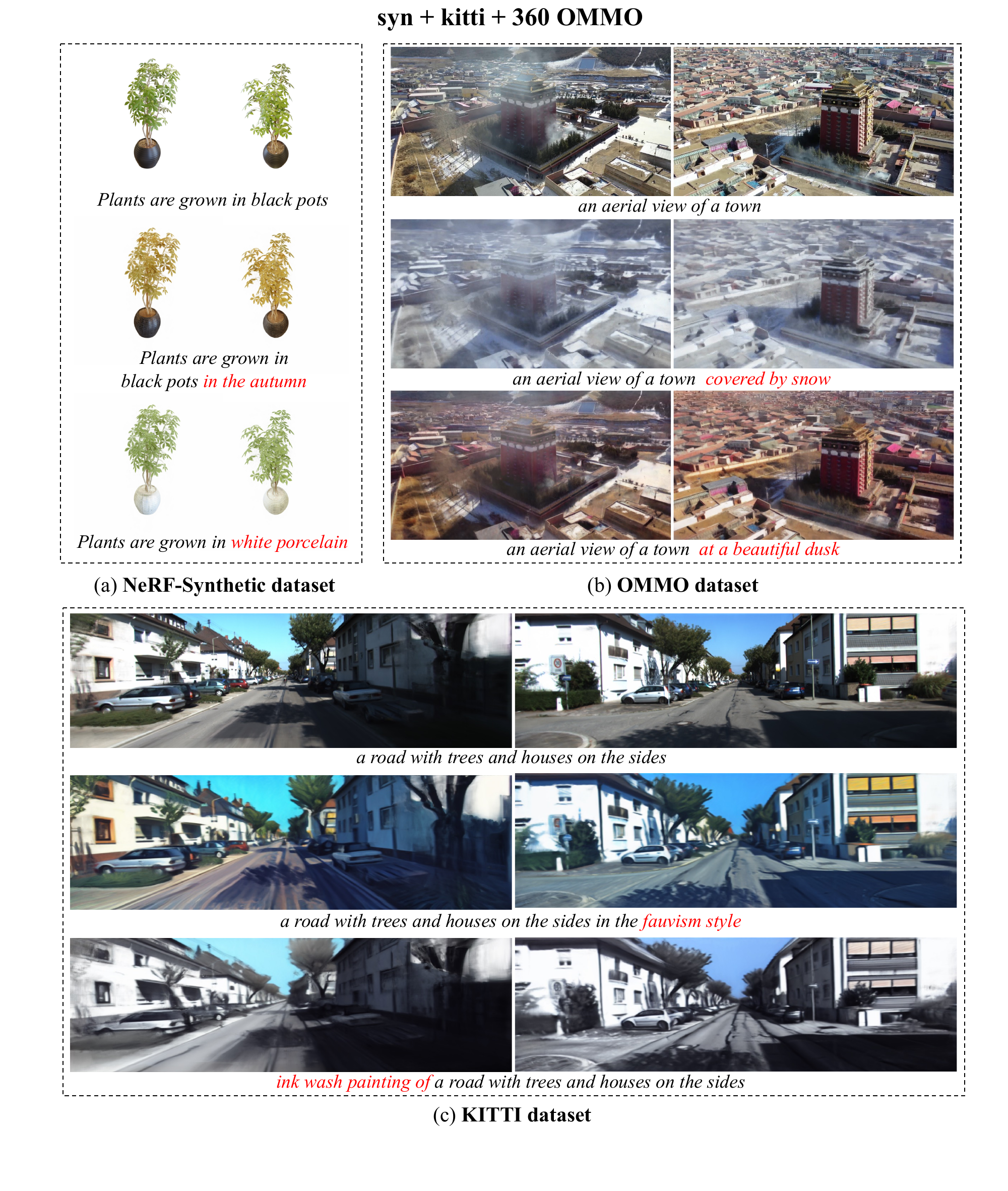}}
        \caption{\textbf{More visual results on NeRF-Synthetic, OMMO, and KITTI datasets}. The NeRF-Synthetic dataset~\citep{mildenhall2020nerf} is synthetic scenes. The OMMO dataset~\citep{lu2023large-ommo} is $360^\circ$ scenes captured by the drone camera. The KITTI dataset~\citep{geiger2013kitti} is the street view scenes captured by the car camera.}
        \label{fig:ommo_kitti}
        \end{center}
    \vspace{-4 ex}
\end{figure*}

\begin{figure*}[htbp]
        \centering
        \begin{center}
        \centerline{\includegraphics[width=0.9\linewidth]{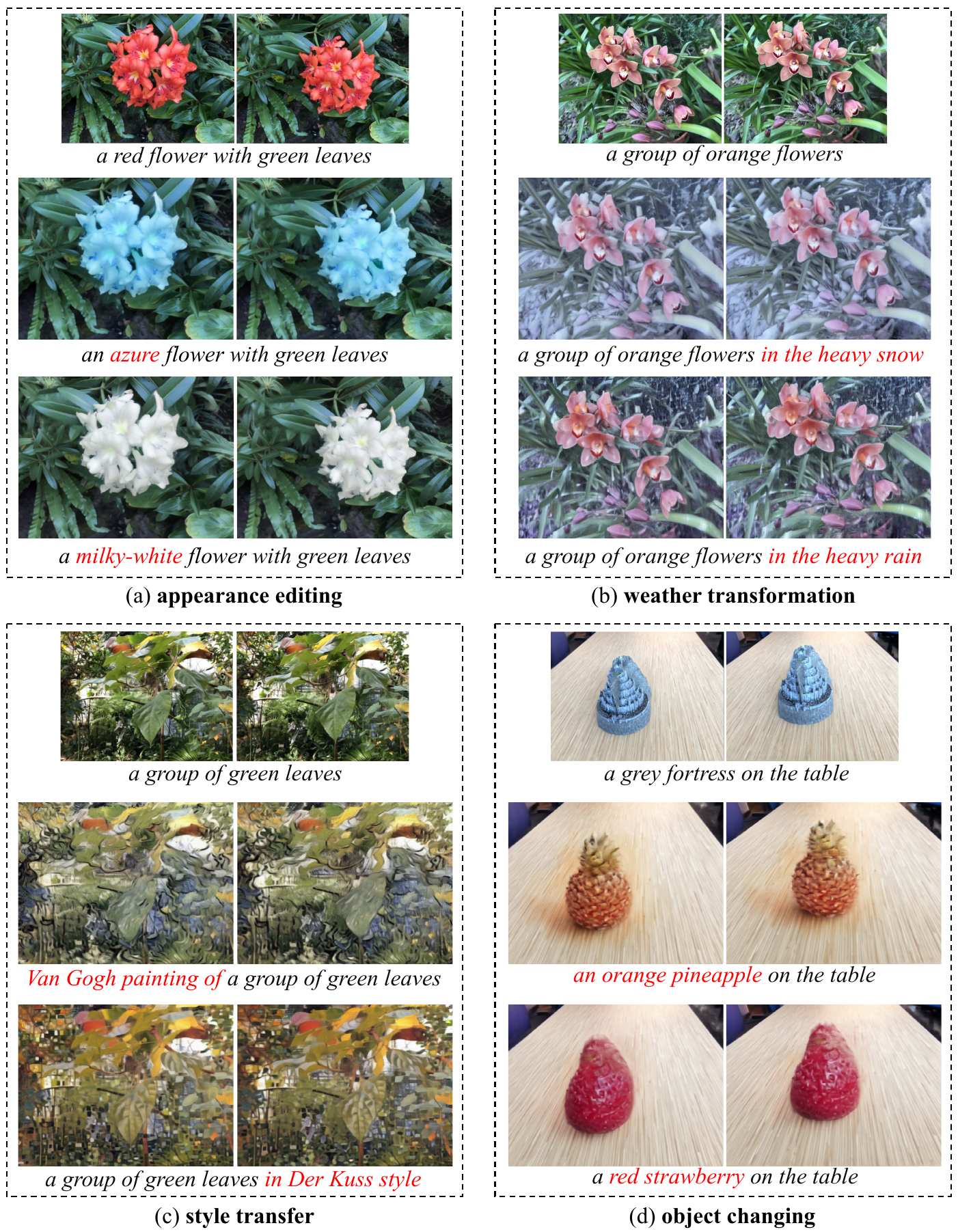}}
        \caption{\textbf{More visual results on LLFF dataset}.}
        \label{fig:more_res_llff}
        \end{center}
    \vspace{-4 ex}
\end{figure*}

\begin{figure*}[htbp]
        \centering
        \begin{center}
        \centerline{\includegraphics[width=\linewidth]{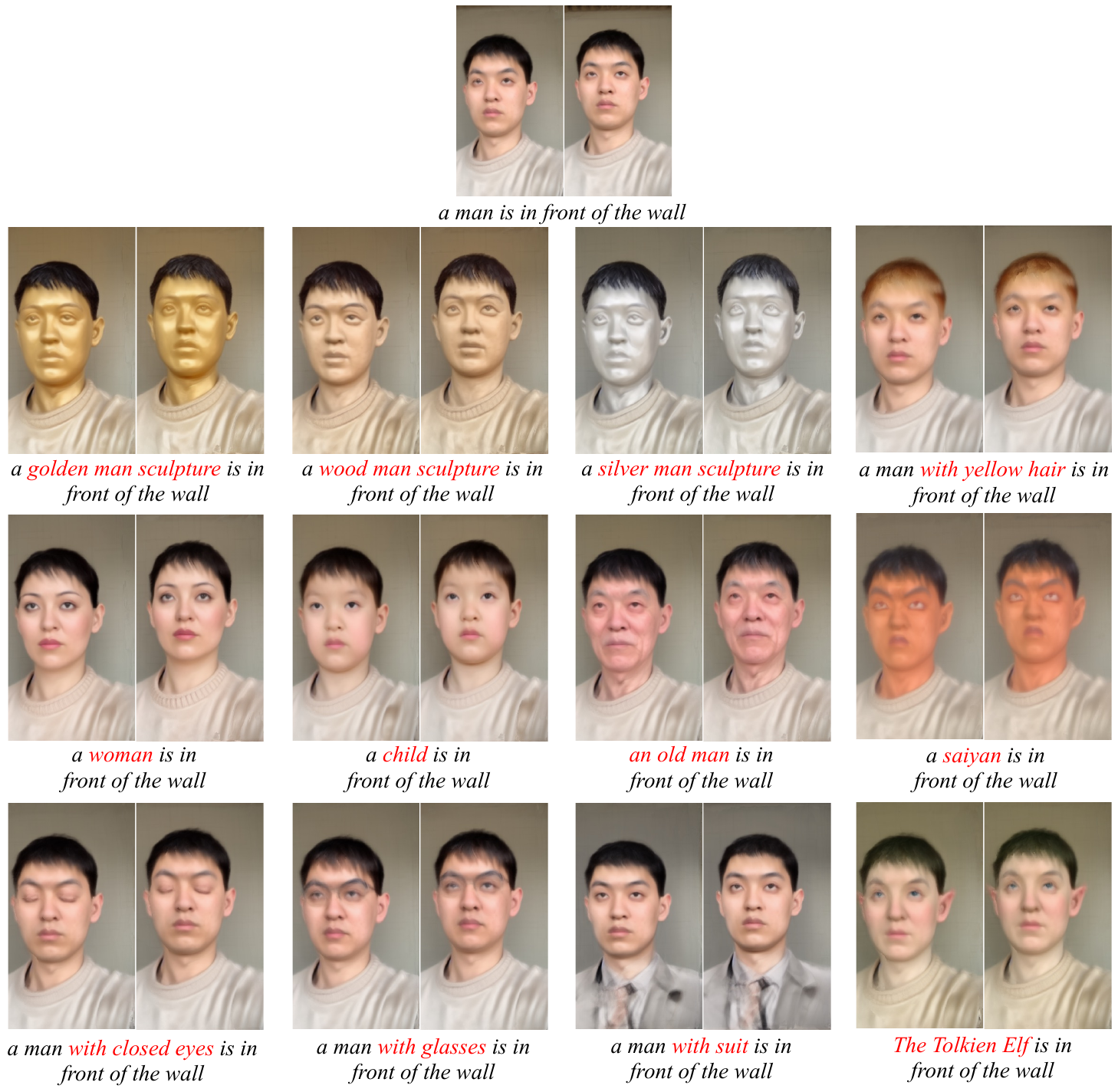}}
        \caption{\textbf{More visual results on portrait editing}.}
        \label{fig:more_res_fang}
        \end{center}
    \vspace{-4 ex}
\end{figure*}

\begin{figure*}[htbp]
        \centering
        \begin{center}
       \centerline{\includegraphics[width=0.7\linewidth]{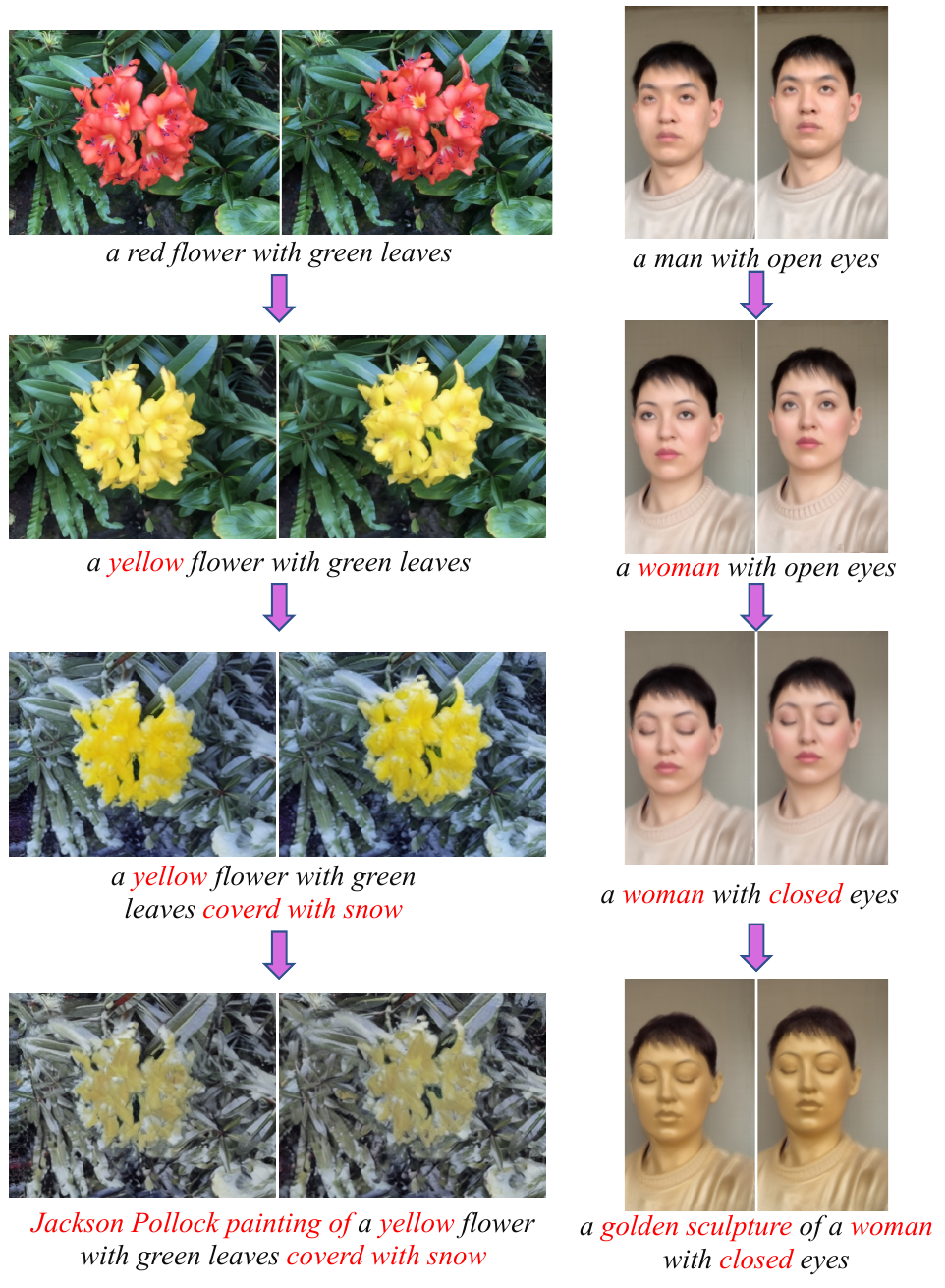}}
        \caption{\textbf{Cascade editing results}. We begin by applying an edit to the original scene and then generate a new scene with the desired editing effect using DN2N. We repeat this process by applying another edit to the new scene. In this figure, we showcase the consecutive results of three cascaded edits.}
        \label{fig:cascade}
        \end{center}
\end{figure*}

In this section, we present a collection of additional editing results that showcase the remarkable editing capabilities and effects of our method. The results are shown as 
Figure~\ref{fig:ommo_kitti}
Figure~\ref{fig:more_res_llff}, Figure~\ref{fig:more_res_fang}, and
in Figure~\ref{fig:cascade}.

\end{document}